\documentclass[twoside,11pt]{article}

\usepackage[accepted]{melba}

\usepackage{amsmath,amsfonts}

\usepackage{color}
\usepackage[table]{xcolor} 
\usepackage{tabularx} 
\usepackage{gensymb} 
\usepackage{amssymb} 
\usepackage{bm} 
\usepackage{booktabs} 
\usepackage{amsmath}
\usepackage{mathtools}
\usepackage{multirow}
\usepackage{subfig}
\usepackage{tabu}

\usepackage{xcolor}
\usepackage{colortbl}
\usepackage{arydshln}

\usepackage{amsmath}

\DeclareMathOperator*{\argmin}{arg\,min}

\renewcommand{\vec}[1]{\mathbf{#1}}

\newcommand{\target}[1]{#1}
\newcommand{\prediction}[1]{\hat{#1}}
\newcommand{\coordinate}{\vec{x}}
\newcommand{\covariance}{\mathbf{\Sigma}}
\newcommand{\diagsigma}{\mathbf{\Sigma}^*}
\newcommand{\rotation}{\mathbf{R}}
\newcommand{\targetcoordinate}{\target{\coordinate}}
\newcommand{\predictioncoordinate}{\prediction{\coordinate}}
\newcommand{\targetcovariance}{\target{\covariance}}
\newcommand{\predictioncovariance}{\prediction{\covariance}}
\newcommand{\heatmap}{h}

\newcommand{\targetheatmap}{\target{\heatmap}}
\newcommand{\predictionheatmap}{\prediction{\heatmap}}

\newcommand{\sigmamajor}{\sigma^\text{maj}}
\newcommand{\sigmaminor}{\sigma^\text{min}}
\newcommand{\targetsigmamajor}{\target{\sigma}\vphantom{\sigma}^\text{maj}}
\newcommand{\targetsigmaminor}{\target{\sigma}\vphantom{\sigma}^\text{min}}

\newcommand{\netdirallfit}{\mbox{$\covariance$-target,$\hat{\covariance}$-fit}}

\newcommand{\fixediso}{\mbox{$\sigma$=3,$\hat{\sigma}$-fit}}
\newcommand{\fixedaniso}{\mbox{$\sigma$=3,$\hat{\covariance}$-fit}}

\newcommand{\isoaniso}{\mbox{$\sigma$-target,$\hat{\covariance}$-fit}}

\newcommand{\anisoaniso}{\mbox{$\covariance$-target,$\hat{\covariance}$-fit}}

\newcommand{\mcdpointfit}{\mbox{\text{MCD-}$\predictioncoordinate$}}
\newcommand{\mcdhmean}{\mbox{\text{MCD-}$\predictionheatmap$}}

\newcommand{\mcdhmeananiso}{\mbox{\text{MCD-}$\predictionheatmap$,$\hat{\covariance}$-fit}}

\definecolor{franzcolor}{rgb}{0.1, 0.8, 0.4}
\definecolor{darkocolor}{rgb}{0.1, 0.4, 0.8}
\definecolor{christiancolor}{rgb}{0.8, 0.4, 0.1}

\definecolor{graycolor}{rgb}{0.6, 0.6, 0.6}
\newcommand{\gray}[1]{\textcolor{graycolor}{#1}}

\definecolor{ratiocolor}{rgb}{0.88, 1.0, 1.0}

\definecolor{reviewerOne}{rgb}{0.6, 0.6, 0.0}

\definecolor{reviewerTwo}{rgb}{0.6, 0.0, 0.6}

\definecolor{reviewerThree}{rgb}{0.0, 0.6, 0.6}

\definecolor{todocolor}{rgb}{1.0, 0.6, 0.0}

\definecolor{red}{rgb}{1.0, 0.0, 0.0}

\makeatletter
\def\thickhline{%
  \noalign{\ifnum0=`}\fi\hrule \@height \thickarrayrulewidth \futurelet
   \reserved@a\@xthickhline}
\def\@xthickhline{\ifx\reserved@a\thickhline
               \vskip\doublerulesep
               \vskip-\thickarrayrulewidth
             \fi
      \ifnum0=`{\fi}}
\makeatother
\newlength{\thickarrayrulewidth}
\usepackage{array}
\makeatletter
\g@addto@macro{\endtabular}{\rowfont{}}
\makeatother
\newcommand{\rowfonttype}{}
\newcommand{\rowfont}[1]{
   \gdef\rowfonttype{#1}#1%
}
\newcolumntype{L}{>{\rowfonttype}l}

\melbaheading{014}{https://www.melba-journal.org/article/27656}{2021}{1-27}{01/2021}{09/2021}{Franz Thaler, Christian Payer, Martin Urschler and Darko \v{S}tern}{Uncertainty for Safe Utilization of Machine Learning in Medical Imaging (UNSURE) 2020}{Christian Baumgartner, Adrian Dalca, Carole Sudre, Ryutaro Tanno, Sandy Wells}

\ShortHeadings{Modeling Annotation Uncertainty with Gaussian Heatmaps}{Thaler, Payer, Urschler and \v{S}tern}
\firstpageno{1}

\title{Modeling Annotation Uncertainty with\\Gaussian Heatmaps in Landmark Localization}

\author{\renewcommand*{\thefootnote}{\arabic{footnote}} \\ 
    \name Franz Thaler$^{*}$ \email franz.thaler@medunigraz.at \\
	\addr Gottfried Schatz Research Center: Biophysics, Medical University of Graz, Austria
	\AND
	\name Christian Payer$^{*}$ \email christian.payer@icg.tugraz.at \\
	\addr Institute of Computer Graphics and Vision, Graz University of Technology, Austria
	\AND
	\name Martin Urschler \email martin.urschler@auckland.ac.nz \\
	\addr School of Computer Science, The University of Auckland, New Zealand
	\AND
	\name Darko \v{S}tern \email darko.stern@medunigraz.at \\
	\addr Gottfried Schatz Research Center: Biophysics, Medical University of Graz, Austria
}

\begin{document}

\maketitle

\renewcommand*{\thefootnote}{\fnsymbol{footnote}}
\footnotetext[1]{contributed equally}

\renewcommand*{\thefootnote}{\arabic{footnote}}

\begin{abstract}
In landmark localization, due to ambiguities in defining their exact position, landmark annotations may suffer from large observer variabilities, which result in uncertain annotations.
To model the annotation ambiguities of the training dataset, we propose to learn anisotropic Gaussian parameters modeling the shape of the target heatmap during optimization.
Furthermore, our method models the prediction uncertainty of individual samples by fitting anisotropic Gaussian functions to the predicted heatmaps during inference.
Besides state-of-the-art results, our experiments on datasets of hand radiographs and lateral cephalograms also show that Gaussian functions are correlated with both localization accuracy and observer variability.
As a final experiment, we show the importance of integrating the uncertainty into decision making by measuring the influence of the predicted location uncertainty on the classification of anatomical abnormalities in lateral cephalograms.
\end{abstract}

\begin{keywords}
  landmark localization, uncertainty estimation, heatmap regression. 
\end{keywords}

\section{Introduction}
Anatomical landmark localization is an important topic in medical image analysis, e.g., as a preprocessing step for segmentation~\citep{Beichel2005,Heimann2009, Stern2011}, registration~\citep{Johnson2002,Urschler2006}, as well as for deriving surgical or diagnostic measures  like the location of bones in the hip~\citep{Bier2019}, curvature of the spine~\citep{Vrtovec2009}, bone age estimation~\citep{Stern2019a}, or misalignment of teeth or the jaw~\citep{Wang2016}.
Unfortunately, anatomical landmarks can be difficult to define unambiguously, especially for landmarks that do not lie on distinct anatomical structures like the tip of the incisor, but on smooth edges like the tip of the chin.
Ambiguity in defined landmark positions leads to difficulty in annotation which is reflected not only in a large inter-observer variability between different annotators, but also in a large intra-observer variability that is dependent on the daily constitution of a single annotator.
Due to this uncertainty, a machine learning predictor that has been trained on potentially ambiguous annotations should model landmarks not only as single locations, but rather as distributions over possible locations.
Taking the inconsistencies introduced by intra- and inter-observer variability of the annotators into account during decision making will lead to a potential improvement in clinical workflow and patient treatment~\citep{Lindner2016}.
Furthermore, estimating the uncertainty that reflects the annotation ambiguities helps interpreting the output of machine learning based prediction software~\citep{Gal2016}, which is especially useful in the medical imaging domain~\citep{Wang2019,Nair2020,Wickstrom2020} where explainability is crucial~\citep{Gunning2019}.


\begin{figure}[t]
\centering
\includegraphics[width=0.8\textwidth]{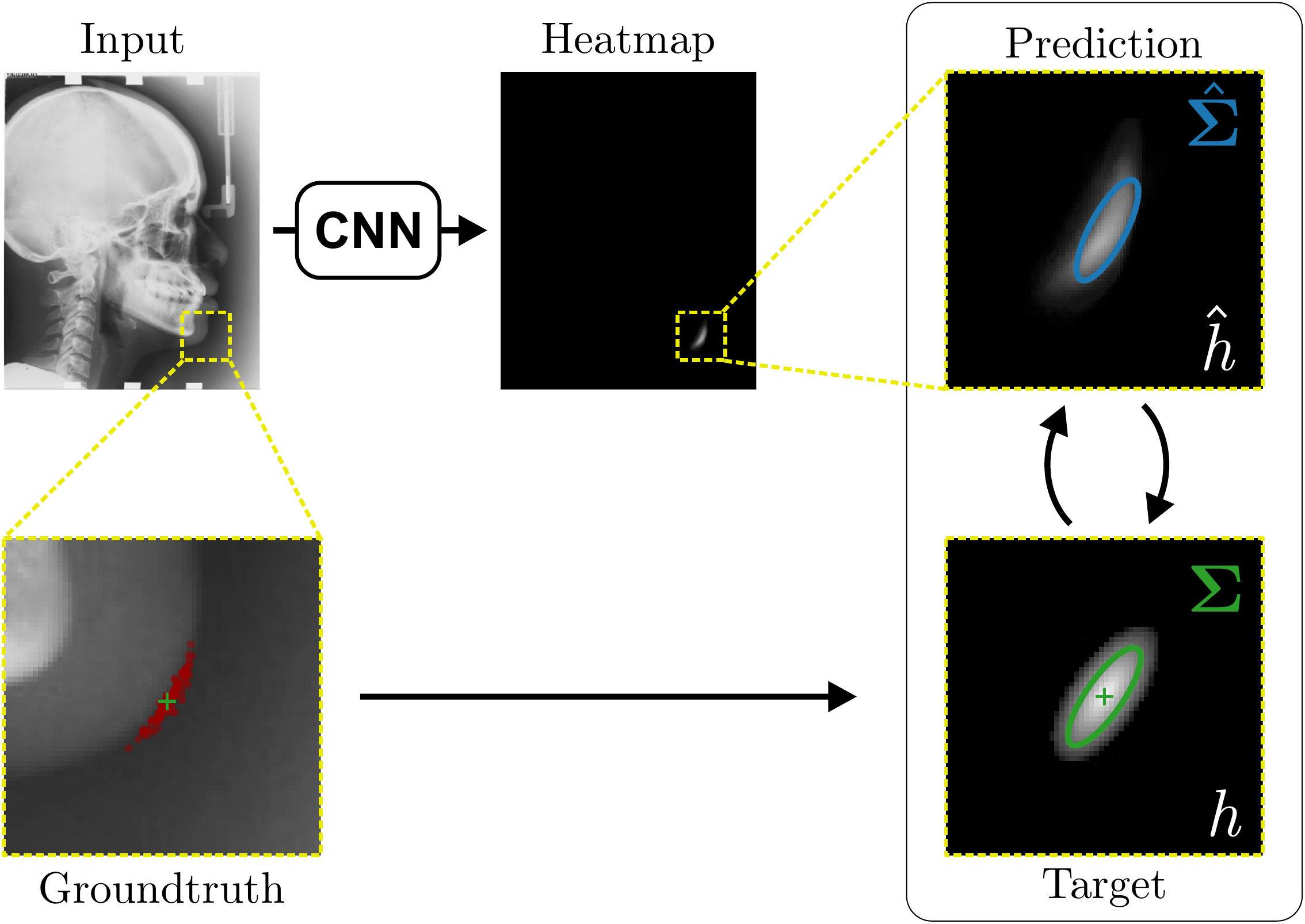}
\caption{
Schematic representation of our proposed method to model annotation ambiguities.
Although annotation ambiguities (red dots) are not observable from a single groundtruth annotation (green cross), they are present in the entirety of the annotated \mbox{images}.
Landmark ambiguities are modeled using a Gaussian function with covariance $\targetcovariance$ that is learned during training and represents a \textit{dataset-based annotation uncertainty}. 
The predicted heatmap $\predictionheatmap$ is fitted with covariance $\predictioncovariance$ to represent the predicted landmark locations as a distribution, modeling a \textit{sample-based annotation uncertainty}.
}
\label{fig:overview}
\end{figure}

\subsection{Contributions}

Differently from state-of-the-art methods that use heatmap based regression for predicting single landmark locations, in this work we show that heatmaps can also be used to estimate the landmark uncertainty caused by annotation ambiguities.
To this end, we first adapt the heatmap regression framework to allow optimization of not only the model parameters of the network but also the shape parameters of the target heatmaps. 
By doing so, we model the uncertainty based on annotation ambiguities of the entire training dataset, representing a homoscedastic aleatoric uncertainty that is the same for each image in the dataset.
We name this uncertainty \textit{dataset-based annotation uncertainty}.
Second, during inference, a Gaussian function is fitted to the predicted heatmap to not only predict the most probable landmark location but its distribution, i.e., a heteroscedastic aleatoric uncertainty that is specific for each image.
We name this uncertainty \textit{sample-based annotation uncertainty}.
We evaluate our approach on two datasets, i.e., on left hand radiographs and lateral cephalograms, showing that the uncertainties correlate with the magnitude and direction of the average localization error as well as the inter-observer variabilities.

This work extends our MICCAI-UNSURE 2020 workshop paper~\citep{payer2020uncertainty} in the following:
(1)~We improved the representativeness of the inter-observer variability used in our experiments by extending the annotation of the cephalogram dataset with additional annotations from nine observers.
(2)~We performed an in-depth analysis of our proposed method for estimating the \textit{dataset-based} and \textit{sample-based annotation uncertainty} on the newly annotated dataset.
(3)~We introduced a new experiment showing the importance of utilizing landmark location uncertainty into decision making with an example of measuring anatomical abnormalities in lateral cephalograms.
(4)~Finally, we included more details of the method, and extended our literature review and discussion section.

\section{Related Work}

\subsection{Landmark Localization}

Due to the variable visual appearance 
of the anatomy defining a landmark position in images, modern methods for landmark localization are based on convolutional neural networks (CNNs) because of their superior image feature extraction capabilities as compared to other machine learning techniques~\citep{Toshev2014,Tompson2014,Wei2016,Payer2016,Newell2016,Yang2017,Payer2019,Bier2019}. 
While earlier work that applied CNNs for human pose estmation -- a task similar to anatomical landmark localization -- uses CNNs for directly regressing landmark coordinates \citep{Toshev2014}, more recent methods for human pose estimation apply CNNs with the heatmap regression framework \citep{Tompson2014,Wei2016,Newell2016}.
In the heatmap regression framework, the network is trained not to directly predict the landmark coordinates, but a heatmap which is generated as an image of a Gaussian function centered at the coordinate of the landmark. 
Thus, in this image-to-image mapping, the network learns to generate high responses on locations close to the target landmarks, while responses on wrong locations are being suppressed.

In our previous work~\citep{Payer2016}, we introduced the CNN-based heatmap regression framework to anatomical landmark localization, while additionally proposing a CNN architecture called SpatialConfiguration-Net (SCN), which integrates spatial information of landmarks directly into an end-to-end trained, fully convolutional network.
Building upon our approach, \cite{Yang2017} used the heatmap regression framework to generate predictions for landmarks with missing responses and incorporated an additional postprocessing step to remove false-positive responses.
\cite{Bier2018} used heatmap regression for localizing landmarks on radiographs of the hips.
To cope with the small number of images of their training dataset, they implement a sophisticated image synthesis framework based on projecting three-dimensional volumes to two-dimensional images and pretrain their networks with these synthetic images.
\cite{Mader2019} adapted the heatmap regression networks for predicting more than a hundred landmarks on a dataset of spinal CT volumes.
However, due to memory restrictions resulting from a large number of landmarks, they do not incorporate upsampling layers, thus reducing the accuracy of their heatmap regression outputs.
Other recent methods use a coarse-to-fine approach by combining global and local image information with an attention-guided mechanism \citep{Zhong2019}, or with multiplying intermediate network outputs to restrict local responses to their global position \citep{Chen2019}. 

In all aforementioned methods, the size of the target heatmaps, i.e., the variance of the Gaussian function, is fixed for each landmark and needs to be predefined before network training.
Although our previous work in \cite{Payer2019} indicated that learning the parameters of an isotropic target Gaussian function for each landmark heatmap independently increases prediction accuracy, we did not provide any interpretation of the estimated heatmap size.
Moreover, while there exist work that fits isotropic Gaussian functions to the predicted heatmap outputs for obtaining a more robust point prediction, e.g., \cite{Liu2019}, most methods only take the single point-wise coordinate prediction as a final output and are not considering the remaining information in the heatmap prediction.
Thus, to the best of our knowledge, none of the methods has investigated how the target Gaussian functions or the Gaussian function fitted to the heatmap prediction correlates with the uncertainty of landmark localization.

\subsection{Uncertainty Estimation}
Uncertainty estimation in medical imaging has been widely adopted and used in segmentation, e.g., for segmenting multiple sclerosis lesions~\citep{Nair2020}, the whole-brain~\citep{roy2018inherent}, the fetal brain and brain tumors~\citep{wang2019aleatoric}, the carotid artery~\citep{camarasa2020quantitative} as well as for ischemic stroke lesion and vessel extraction from retinal images~\citep{kwon2020uncertainty}. 
Furthermore, uncertainty estimation was applied for disease detection using a diabetic retinopathy dataset~\citep{leibig2017leveraging,ayhan2018test}, for classification of skin lesions~\citep{mobiny2019risk} and for probabilistic classifier-based image registration~\citep{sedghi2019probabilistic}.
A number of uncertainty metrics have been proposed in literature, which can be categorized into model (\textit{epistemic}) and data (\textit{aleatoric}) uncertainty~\citep{kendall2017uncertainties}.
While epistemic uncertainty represents the prediction confidence of the model for an unseen sample based on its similarity to training data samples, aleatoric uncertainty is caused by the observation noise inherent to the dataset. 

Epistemic uncertainty for neural networks involves the use of probabilistic models, where the prediction given an image is not deterministic.
This can be achieved with Bayesian Neural Networks (BNN), where weights are not represented as scalars but as distributions.
During training, the parameters of the weight distributions are learned, while during inference, the probabilistic output is generated by combining multiple predictions for different weights sampled from the learned distributions~\citep{blundell2015weight,shridhar2019comprehensive}.
A popular approximation of BNNs is Monte Carlo-Dropout (MCD)~\citep{Gal2016}.
Here, the weight distributions are approximated by reusing Dropout~\citep{Srivastava2014} during inference to retrieve a set of stochastic predictions.
Since MCD is easy to implement and computationally more efficient than BNNs, it is the most common method to model epistemic uncertainty in medical imaging~\citep{leibig2017leveraging,roy2018inherent,mobiny2019risk,wang2019aleatoric,camarasa2020quantitative,kwon2020uncertainty,Nair2020}.

Aleatoric uncertainty for neural networks can be estimated by considering network outputs as distributions rather than point predictions.
For classification tasks, this can be achieved by using not only the most probable class as the prediction, but the distribution represented by the probabilistic network outputs~\citep{kendall2017uncertainties}.
For regression tasks, this can be achieved by adding additional network outputs that model the assumed distribution of the observed annotation noise~\citep{kendall2017uncertainties,kwon2020uncertainty}.
For example, when assuming the annotation noise is normally distributed, the network is trained to not only predict the mean but also the variance of the distribution as an additional network output.
Aleatoric uncertainty has only recently been investigated in landmark localization for computer vision tasks like pose estimation \citep{Chen2019iccv, Gundavarapu2019_CVPR_Workshops}.
Nevertheless, it was only done in the context of occluded landmarks.
Due to large inter- and intra-observer variabilities \citep{chotzoglou2019exploring,jungo2018effect}, the dominant source of aleatoric uncertainty in the medical domain is annotation ambiguity. 
To the best of our knowledge, there is no previous work that shows how to evaluate annotation uncertainty in medical images from noisy data annotation.

\section{Uncertainty Estimation in Landmark Localization}
\label{sec:uncertainty estimation}


\subsection{Heatmap Regression for Dataset-Based Uncertainty}
\label{sec:heatmap_regression}

In landmark localization the task is to use an input image
${I: \Omega_I \rightarrow \mathbb{R}}$ with $\Omega_I \subset \mathbb{R}^2$
to determine the coordinates $\coordinate_i \in \Omega_I$ of $N$ specific landmarks $L_i$ within the extent $\Omega_I$ of the image $I$.
In the heatmap regression framework, a CNN with parameters $\vec{w}$ is
set up
to predict $N$ heatmaps
$\predictionheatmap_i: \Omega_I \rightarrow \mathbb{R}$, representing the pseudo-probability of the location of landmark $L_i$.
As has been done in several other works (e.g., \cite{Tompson2014, Newell2016, Payer2016, Zhong2019}), a target heatmap
${\targetheatmap_i: \Omega_I \rightarrow \mathbb{R}}$
for landmark $L_i$ with coordinate $\targetcoordinate_i$ can be represented as an isotropic two-dimensional Gaussian function, i.e.,
\begin{equation}
\targetheatmap_i(\vec{x}; \sigma_i) = \frac{\gamma}{2 \pi \sigma_i^2} \exp{\left(-\frac{\lVert \vec{x}-\targetcoordinate_i\rVert_2^2}{2\sigma_i^2}\right)}.
\label{eq:isotropic_gauss}
\end{equation}
The mean of the Gaussian is set to the target landmark's coordinate
${\targetcoordinate_i\in\Omega_I}$,
while the size of the isotropic Gaussian is defined by $\sigma_i$.
A scaling factor $\gamma$ is used to avoid numerical instabilities during training.
Fully convolutional networks that predict $N$ heatmaps $\predictionheatmap_i$ can then be trained with a pixel-wise $L_2$ loss to minimize the difference between predicted heatmaps $\predictionheatmap_i$ and target heatmaps $\targetheatmap_i$ for all landmarks $L_i$ by updating the network parameters $\vec{w}$, i.e.,
\begin{equation}
\argmin_{\vec{w}} \sum_{i=1}^N \sum_{\vec{x} \in \Omega_I} \lVert \predictionheatmap_i(\vec{x}; \vec{w}) - \targetheatmap_i(\vec{x}; \sigma_i) \rVert_2^2.
\label{eq:l2_loss}
\end{equation}

Crucial hyperparameters when optimizing the CNN are the sizes $\sigma_i$ of the Gaussian heatmaps, since suboptimal $\sigma_i$ lead to either reduced prediction accuracy or reduced robustness~\citep{Payer2019}. 
All landmark specific $\sigma_i$ can be found via hyperparameter searching techniques (e.g., grid search), however, such approaches are time consuming. 
Therefore, in our previous work \citep{Payer2019}, sigmas which are unique for each landmark are learned directly during network optimization.
Thus, in \cite{Payer2019}, we adapted the pixel-wise loss function from Eq.~\eqref{eq:l2_loss} to
\begin{equation}
\argmin_{\vec{w}, \sigma_1,\dots,\sigma_N} \sum_{i=1}^N \sum_{\vec{x} \in \Omega_I} \lVert \predictionheatmap_i(\vec{x}; \vec{w}) - \targetheatmap_i(\vec{x}; \sigma_i) \rVert_2^2 + \alpha \sum\limits_{i=1}^N \sigma_i^2.
\label{eq:l2_loss_isotropic}
\end{equation}
Since the $\sigma_i$ parameters of the Gaussian functions are also treated as unknowns, we enable learning them in addition to the network parameters~$\vec{w}$.
Thus, the gradient is not only backpropagated through the predicted heatmaps $\predictionheatmap_i$ to update the network parameters $\vec{w}$, but also through the target heatmaps $\targetheatmap_i$ to update the individual target heatmap sizes $\sigma_i$.
The second term in Eq.~\eqref{eq:l2_loss_isotropic} is used as a regularization to prefer smaller sigma values, as otherwise minimizing the difference between $\predictionheatmap_i$ and $\targetheatmap_i$ could lead to the trivial solution where $\sigma_i \rightarrow \infty$ with $\targetheatmap_i \approx 0$.
It is important to note that although the learned target heatmap sizes are different for each landmark, each target heatmap size $\sigma_i$ is the same for all images in the dataset.

As we will show in our experiments, learning $\sigma_i$ values enables the network to encode the prediction uncertainty for each landmark individually. 
The more certain the network is in its prediction of a specific landmark, the smaller the target heatmap size will be.
However, when using isotropic Gaussian functions as in Eq.~\eqref{eq:isotropic_gauss} as the target heatmaps, information of the directional uncertainty cannot be modeled, which would be beneficial for landmarks that are ambiguous only in a specific direction (see Fig.~\ref{fig:overview}).

In this work, we propose to model the target heatmaps with anisotropic Gaussian functions.
A target heatmap is represented as an image of an anisotropic two-dimensional Gaussian function
${\targetheatmap_i(\vec{x};\targetcovariance_i) : \Omega_I \rightarrow \mathbb{R}}$, i.e.,
\begin{equation}
\targetheatmap_i(\vec{x};\targetcovariance_i) = \frac{\gamma}{2 \pi\sqrt{\begin{vmatrix}\targetcovariance_i\end{vmatrix}}} \exp{\left(-\frac{1}{2} (\vec{x}-\targetcoordinate_i)^T\targetcovariance\vphantom{\covariance}_i^{-1}(\vec{x}-\targetcoordinate_i)\right)}.
\label{eq:heatmap:gauss}
\end{equation}
Differently to Eq.~\eqref{eq:isotropic_gauss}, the shape of the Gaussian is not defined by only a single $\sigma_i$ value but by a full covariance matrix $\targetcovariance_i$.
This way, the target heatmaps are able to have an orientation and varying extents in different directions.
To separately analyze the orientation and the extent, the covariance matrix $\targetcovariance_i$ can be decomposed into
\begin{equation}
\targetcovariance_i = \rotation_i \, \diagsigma_i \, \rotation^T_i
\end{equation}
with
\begin{equation}
\rotation_i = \begin{bmatrix*}[r]\cos \theta_i &\; -\sin \theta_i \\\sin \theta_i &\; \cos \theta_i \\\end{bmatrix*}
\qquad
\text{and}
\qquad
\diagsigma_i = \begin{bmatrix*}[l](\targetsigmamajor_i)^2 &\; 0 \\0 &\; (\targetsigmaminor_i)^2 \\\end{bmatrix*},
\label{eq:heatmap:cov}
\end{equation}
where $\theta_i$ represents the angle of the Gaussian function's major axis, and $\sigmamajor_i$ and $\sigmaminor_i$ represent its extent in major and minor axis, respectively.

The final loss function for simultaneously regressing $N$ heatmaps is defined as 
\begin{equation}
\argmin_{\vec{w},\covariance_1,\dots,\covariance_N}\sum\limits_{i=1}^N \sum_{\vec{x} \in \Omega_I} \lVert \predictionheatmap_i(\vec{x};\vec{w}) -  \targetheatmap_i(\vec{x};\covariance_i) \rVert_2^2 + \alpha \sum\limits_{i=1}^N \sigmamajor_i \sigmaminor_i. 
\label{eq:heatmap:opt}
\end{equation}
Same as in Eq.~\eqref{eq:l2_loss_isotropic}, the first term in Eq.~\eqref{eq:heatmap:opt} minimizes the differences between the predicted heatmaps $\predictionheatmap_i$ and the target heatmaps~$\targetheatmap_i$ for all landmarks $L_i$, while we now not only learn the isotropic $\sigma_i$ but all covariance parameters of the Gaussian functions ($\theta_i$, $\sigmamajor_i$, and $\sigmaminor_i$).
Same as before, the second term of Eq.~\eqref{eq:l2_loss_isotropic} is used to prefer smaller Gaussian extents by penalizing $\sigmamajor_i \sigmaminor_i$ with factor~$\alpha$.
Note that while the mean parameters of the Gaussian functions ($\targetcoordinate_i$) are set to the groundtruth landmark locations and differ for each training image, the covariance parameters of the target heatmaps $\targetheatmap_i$ are learned for the whole dataset.

Analyzing Eq.~\eqref{eq:heatmap:opt} in more detail, not only the predicted heatmaps $\predictionheatmap_i$ aim to be close to the target heatmaps $\targetheatmap_i$ 
but also the target heatmaps $\targetheatmap_i$ aim to be close to the predicted heatmaps $\predictionheatmap_i$. 
For each landmark $L_i$, $\predictionheatmap_i$ and $\targetheatmap_i$ receive feedback from each other during training (see Fig.~\ref{fig:overview}).
While the network parameters $\vec{w}$ are updated to better model the shape of the target heatmap $\predictionheatmap_i$, at the same time, the covariance $\targetcovariance_i$ parameters of each target heatmap $\targetheatmap_i$ ($\theta_i$, $\sigmamajor_i$, and $\sigmaminor_i$) are updated to better model the shape of the predicted heatmap $\predictionheatmap_i$.
Note that, while $\targetcovariance_i$ is learned from a single annotation per image, the CNN sees one annotation for each image in the training set.
This allows implicitly modeling the ambiguities of landmark annotations for the whole training dataset representing a \textit{dataset-based annotation uncertainty}.

\subsection{Heatmap Fitting for Sample-Based Uncertainty}
\label{sec:heatmap_fitting}

As the output of heatmap regression networks are $N$ heatmaps that correspond to the $N$ landmarks $L_i$ rather than their coordinates, each landmark's coordinate $\predictioncoordinate_i$ needs to be obtained from the predicted heatmaps $\predictionheatmap_i$.
Previous work often uses the coordinate of the maximum response or the center of mass of the predicted heatmaps as the landmark's coordinate. 
However, this represents only the most probable landmark location, but not the whole distribution of possible landmark locations. 

As the target heatmaps are modeled as Gaussian functions, the predicted heatmaps are also expected to be Gaussian functions, see Eq.~\eqref{eq:heatmap:gauss}.
Hence, the distribution of possible landmark locations is captured by fitting Gaussian functions to the predicted heatmaps~$\predictionheatmap$.
We use a robust least squares curve fitting method \citep{Branch1999} to fit the Gaussian function (see Eq.~\eqref{eq:heatmap:gauss}) to the predicted heatmaps $\predictionheatmap_i$ initialized to the respective maximum and obtain the fitted Gaussian parameters $\predictioncoordinate_i$ and $\predictioncovariance_i$ with $i=\{1,...,N\}$.
As the fitted Gaussian function models the distribution of possible locations, the parameters represent a directional \textit{sample-based annotation uncertainty} of the landmark location during inference.

\section{Experimental Setup}
\label{sec:setup}

\subsection{Networks}
\paragraph{SpatialConfiguration-Net for Anatomical Landmark Localization:}
Our proposed method to learn anisotropic Gaussian heatmaps can be combined with any image-to-image based network architecture.
In this work, the \mbox{SpatialConfiguration-Net} (SCN) by~\cite{Payer2019} is used, due to its state-of-the-art performance. 
We implement learning of Gaussian functions with arbitrary covariance matrices and use the SCN with the default parameters trained for 40,000 iterations.
Furthermore, training data augmentation is employed, i.e., random intensity shift and scale as well as translation, scaling, rotation, and elastic deformation.
We use $L_2$ weight regularization with a factor of $\lambda=0.001$.
In Eq.~\eqref{eq:heatmap:gauss} we set $\gamma = 100$;
in Eq.~\eqref{eq:heatmap:opt}, we initialize $\sigmamajor_i=\sigmaminor_i=3$, $\theta_i=0$, and set $\alpha = 5$.

\paragraph{Monte Carlo-Dropout to Estimate Label Distributions:}
To compare to a method for estimating uncertainty, we employ Monte Carlo-Dropout (MCD) by~\cite{Gal2016} to predict a set of $K=20$ different heatmaps $\prediction{H}_i = \{\predictionheatmap_i^{(1)}, \dots, \predictionheatmap_i^{(K)} \}$ per landmark~$i$.
Two MCD-based approaches are applied to determine the landmark prediction $\predictioncoordinate_i$ and the covariance parameters $\predictioncovariance_i$:
\\
(1) $\mcdpointfit$,max: 
In a majority of methods using MCD (Leibig et al., 2017; Roy et al., 2018; Camarasa et al., 2020; Kwon et al., 2020;  Nair et al., 2020), the prediction and the uncertainty are computed as the mean and the variance of multiple forward predictions, respectively.
Following these approaches, first, the maximum of each heatmap $\predictionheatmap^{(k)}_i$ is extracted to retrieve a set of stochastic predictions $\prediction{\vec{X}}_i = \{\predictioncoordinate^{(1)}_i, \dots, \predictioncoordinate^{(K)}_i \}$.
Then, the predicted location is computed as the expected value $E$, i.e., the mean $\predictioncoordinate_i^{\text{MCD}} = E[\prediction{\vec{X}}_i]$, and the uncertainty is computed as the covariance
$\predictioncovariance_i^{\text{MCD}} = E[(\prediction{\vec{X}}_i - E[\prediction{\vec{X}}_i])(\prediction{\vec{X}}_i - E[\prediction{\vec{X}}_i])^\top]$
of the set of landmark predictions $\prediction{\vec{X}}_i$.
\\
(2) \mcdhmeananiso:
Since the previous approach $\mcdpointfit$,max leads to an underestimation of the uncertainty when the individual landmark predictions are very close to one another, in this approach, we combine MCD with the uncertainty measure that incorporates the shape of the heatmap, as described in Sec.~3.2.
Thus, in this approach MCD is only used to generate a set of stochastic heatmap predictions, i.e., $\prediction{H}_i = \{\predictionheatmap_i^{(1)}, \dots, \predictionheatmap_i^{(K)} \}$, from which a more robust heatmap prediction is calculated, i.e., $\predictionheatmap_i^{\text{MCD}} = E(\prediction{H}_i)$.
Then, the landmark prediction and uncertainty are estimated by robustly fitting a Gaussian function to $\predictionheatmap_i^{\text{MCD}}$.

\subsection{Datasets}
The proposed method is evaluated on two publicly available datasets for anatomical landmark localization.
The first dataset consists of 895 radiographs of left hands of the Digital Hand Atlas Database System~\citep{Gertych2007,Zhang2009} with 37 annotated landmarks per image. 
Each image has been annotated by one of three experts.
The second dataset consists of 400 lateral cephalograms with 19 annotated landmarks per image.
This dataset has been used for the ISBI 2015 Cephalometric X-ray Image Analysis Challenge~\citep{Wang2016}, in which each image has been annotated by both a senior and a junior radiologist. 
%

\paragraph{Additional Annotations:}
To assess the inter-observer variability and compare it with the predicted uncertainty, additional annotations for the dataset of cephalograms were obtained.
100 images were sampled randomly for which nine annotators experienced in medical image analysis labeled five representative landmarks of varying difficulty and ambiguity.
The landmarks were chosen such that they cover a wide range of properties reflected by the shape and size of their expected distributions (see Fig.~\ref{fig:selected_landmarks}).
Including the preexisting annotations of the senior and junior radiologist, this gives us a total of 11 annotations for the five selected landmarks\footnote{The code and additional annotations are available at:\\ \href{https://github.com/christianpayer/MedicalDataAugmentationTool-HeatmapUncertainty}{https://github.com/christianpayer/MedicalDataAugmentationTool-HeatmapUncertainty}}. 
To estimate the distribution of the inter-observer variability, a Gaussian distribution was fitted to the 11 available annotations per landmark and image to retrieve the distribution parameters $\theta$, $\sigmamajor$ and $\sigmaminor$ (see Eq.~\eqref{eq:heatmap:gauss}).
The inter-observer variability on the whole dataset for a landmark is calculated by taking the average over the distribution parameters for all images.




%

\newcommand{\annotationplotwidth}{0.40\textwidth}

\begin{figure}[t]
\centering
\begin{tabular}{cccc}
\includegraphics[width=\annotationplotwidth]{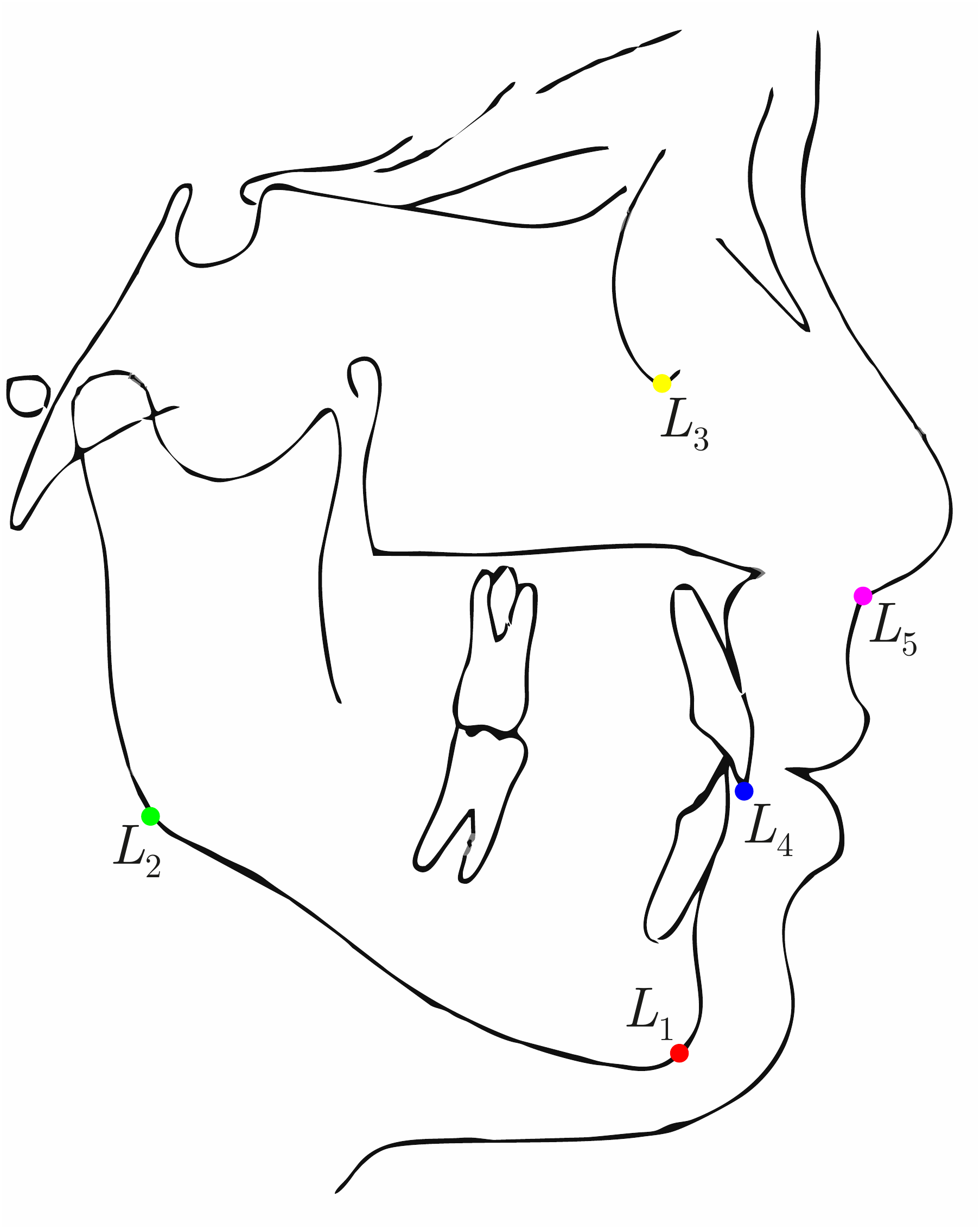} & \includegraphics[width=\annotationplotwidth]{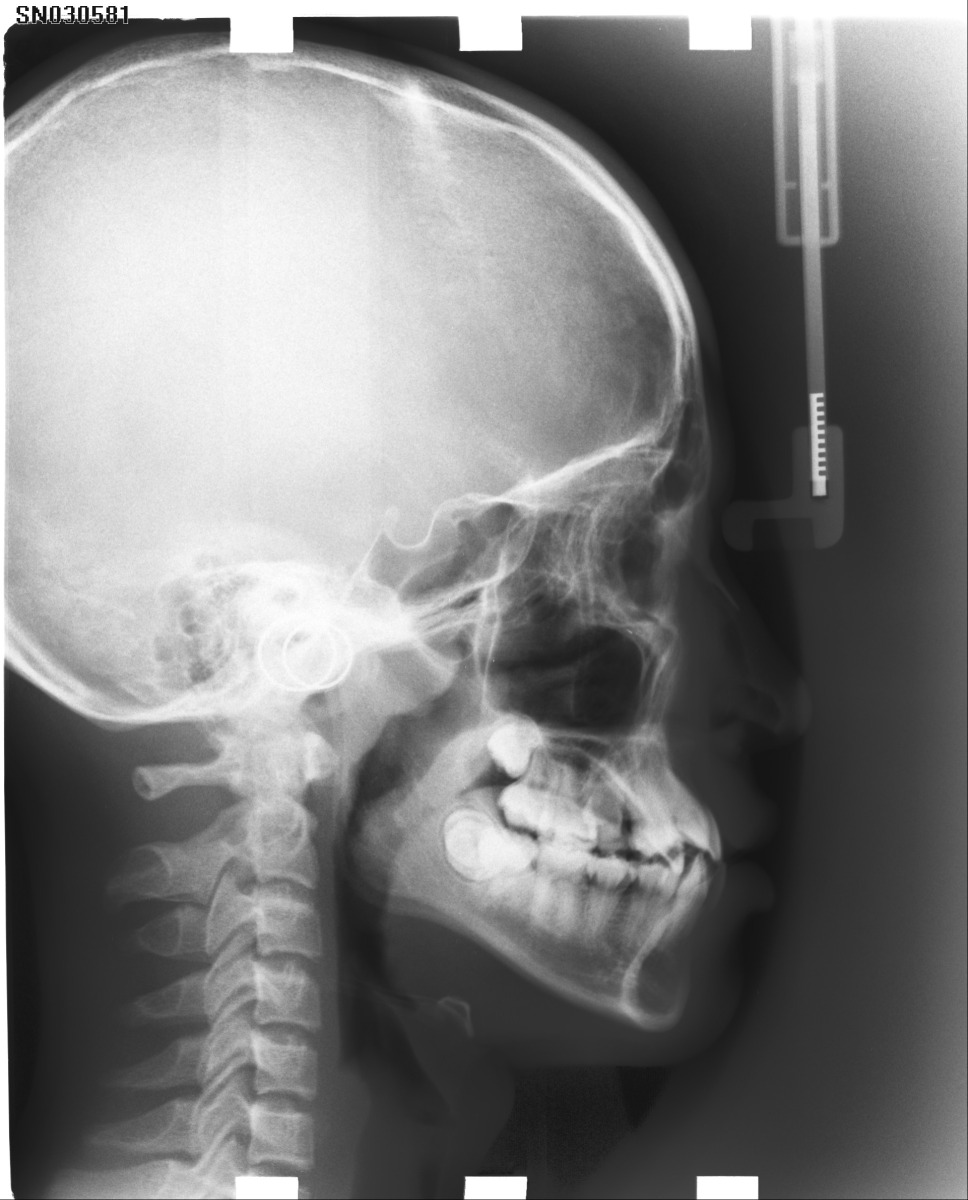} \\
\includegraphics[width=\annotationplotwidth]{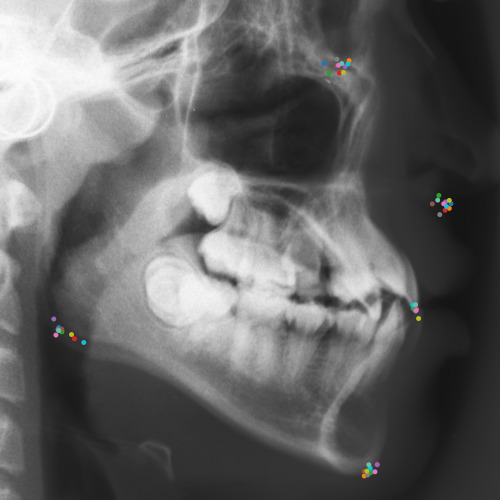} &
\includegraphics[width=\annotationplotwidth]{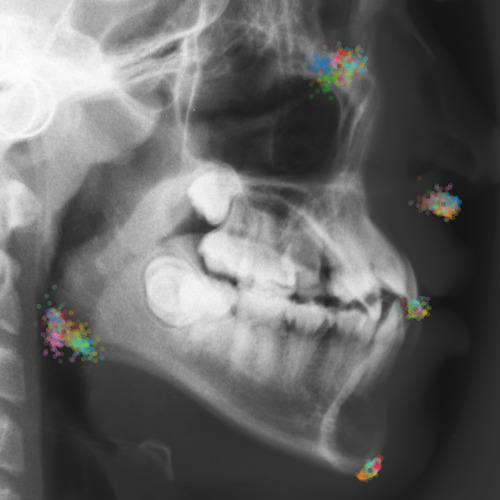}\\
\end{tabular}
\caption{
Schematic of the selected landmarks modified from~\cite{Wang2016} (top left) for which we acquired 9 additional annotations for 100 images and an exemplary image (top right) of the cephalogram dataset.
All annotations for an exemplary image are shown in the bottom left. 
The bottom right shows all annotations of all landmarks in the same image and is generated as follows: 
First, for each image and landmark, the mean location of the 11 annotations, as well as their offsets to this mean are calculated.
Then, for each of the 5 landmarks and 100 images, the 11 annotation offsets are superimposed such that their mean position is centered at the mean position of the corresponding landmark of the exemplary image.
}
\label{fig:selected_landmarks}
\end{figure}


\subsection{Metrics}

For evaluating the landmark localization performance, the point-to-point error is used for the landmarks $i$ in image $j$, which is defined as the Euclidean distance between the target coordinate~$\targetcoordinate^{(j)}_i$ and the predicted coordinate~$\predictioncoordinate^{(j)}_i$, i.e.,
$\text{PE}_i^{(j)} = \lVert \targetcoordinate^{(j)}_i - \predictioncoordinate^{(j)}_i \rVert_2$.
This allows calculation of sets of point-to-point errors, e.g., for all $M$ images of the dataset over all $N$ landmarks, over which the mean and standard deviation can be calculated.
Additionally, we report the success detection rate, which is defined as the percentage of predicted landmarks that are within a certain point-to-point error radius $r$ for all images, i.e.,
$\text{SDR}_r = \frac{100}{M N}\big\vert\{\,(j,i)\;|\;\text{PE}_i^{(j)} \leq r \}\big\vert$.

While in most experiments only a single groundtruth coordinate per image and landmark is available, the groundtruth location in our inter-observer experiment is not only encoded as a single coordinate, but as a distribution. 
Such groundtruth distributions are compared to the distributions predicted by our method by analyzing the parameters of the Gaussian functions.
As a measure of size of the distributions, we use the area encoded by the product of the extent of the major axis and minor axis, i.e., $\sigmamajor \cdot \sigmaminor$.
As a measure of anisotropy, we use their ratio, i.e., $\sigmamajor : \sigmaminor$.
Additionally, to compare the predicted and groundtruth orientation for landmarks with a strong anisotropic distribution, the orientation angle $\theta$ with respect to the x-axis is reported.

\section{Results and Discussion}

In state-of-the-art methods for landmark localization using heatmap regression, only the maximum of the heatmap prediction is used to estimate the landmark location, while the remaining information in the heatmap is neglected.
In this work, we show that the predicted heatmap additionally captures uncertainty information introduced by ambiguities when annotating the training dataset.
Thus, we demonstrate that the parameters of the target heatmap Gaussian function estimated during training are modeling a \textit{dataset-based annotation uncertainty} and the parameters of the Gaussian function fitted to the heatmap prediction models a \textit{sample-based annotation uncertainty} during inference.

First, in Sec.~\ref{sec:localization_experiments} it is shown that our method, which takes the uncertainty of landmark localization into account, outperforms state-of-the-art methods in terms of both accuracy and robustness.
In the following Sec.~\ref{sec:heatmap_uncertainty}, we show that both \textit{dataset-based} and \textit{sample-based annotation uncertainty} correlate with the prediction error, while in Sec.~\ref{sec:inter}, it is demonstrated that uncertainties also model the inter-observer variabilities.
In Sec.~\ref{sec:ablation}, an ablation study is conducted to evaluate different strategies of using Gaussian functions in our method. 
In Sec.~\ref{sec:practical}, we discuss how the inter-observer variability can influence the classification outcome of several clinical measures on lateral cephalograms.


\subsection{Localization Accuracy and Robustness}
\label{sec:localization_experiments}

\begin{table}[t]
\caption{
Quantitative localization results of the hand dataset.
The mean and standard deviation (SD) of the point-to-point error (PE) are given in~mm, the 
success detection rate ($\text{SDR}_r$) shows the percentage of predicted landmarks within a certain PE radius $r$.
Best values are marked in bold.}
\begin{center}
\footnotesize
\begin{tabular}{c|c|cccc}
\hline
\multirow{2}{*}{Method} & $\text{PE}$ (in mm) & \multicolumn{4}{c}{$\text{SDR}_r$ in \%} \\
& mean $\pm$ SD & $r = 2$~mm & $r = 2.5$~mm & $r = 3$~mm & $r = 4$~mm \\
\hline
Lindner et al. (2015) & 0.85 $\pm$ 1.01 & 93.68\% & 96.32\% & 97.69\% & 98.95\% \\
Urschler et al. (2018) & 0.80 $\pm$ 0.93 & 92.19\% & 94.85\% & 96.57\% & 98.46\% \\
Payer et al. (2019) & 0.66 $\pm$ 0.74 & 94.99\% & 96.98\% & 98.12\% & 99.27\% \\
\netdirallfit & \textbf{0.61} $\pm$ 0.67 & \textbf{95.93\%} & \textbf{97.80\%} & \textbf{98.72\%} & \textbf{99.54\%} \\
\hline
\end{tabular}
\end{center}
\label{tbl:results_hand}
\end{table}

\begin{table}[t]
\caption{
Quantitative localization results of the cephalogram dataset for the challenge test sets \textit{Test1} and \textit{Test2}~\citep{Wang2016}, as well as the cross-validation on the junior annotation only (\textit{CV jun.}).
The mean and standard deviation (SD) of the point-to-point error (PE) are given in~mm, the 
success detection rate ($\text{SDR}_r$) shows the percentage of predicted landmarks within a certain PE radius~$r$.
Best values are marked in bold.}
\begin{center}
\footnotesize
\begin{tabular}{cc|c|cccc}
\hline
& \multirow{2}{*}{Method} & $\text{PE}$ (in mm) & \multicolumn{4}{c}{$\text{SDR}_r$ in \%}\\
& & mean $\pm$ SD & $r = 2$~mm & $r = 2.5$~mm & $r = 3$~mm & $r = 4$~mm\\
\hline
\multirow{5}{*}{\rotatebox[origin=c]{90}{\textit{\vphantom{j}{Test1}}}}
& Ibragimov et al. (2014) & 1.84 $\pm$ n/a & 71.72\% & 77.40\% & 81.93\% & 88.04\% \\
& Lindner et al. (2015) & 1.67 $\pm$ n/a & 73.68\% & 80.21\% & 85.19\% & 91.47\% \\
& Zhong et al. (2019) & 1.12 $\pm$ 0.88 & 86.91\% & 91.82\% & 94.88\% & 97.90\%\\
& Chen et al. (2019) & 1.17 $\pm$ n/a & 86.67\% & \textbf{92.67\%} & \textbf{95.54\%} & \textbf{98.53\%}\\
& \netdirallfit & \textbf{1.07} $\pm$ 1.02 & \textbf{87.37\%} & 91.86\% & 94.81\% & 97.79\%\\
\hline
\multirow{5}{*}{\rotatebox[origin=c]{90}{\textit{\vphantom{j}{Test2}}}} 
& Ibragimov et al. (2014) & 2.14 $\pm$ n/a & 62.74\% & 70.47\% & 76.53\% & 85.11\% \\
& Lindner et al. (2015) & 1.92 $\pm$ n/a & 66.11\% & 72.00\% & 77.63\% & 87.42\% \\
& Zhong et al. (2019) & 1.42 $\pm$ 0.84 & \textbf{76.00\%} & \textbf{82.90\%} & \textbf{88.74\%} & 94.32\%\\
& Chen et al. (2019) & 1.48 $\pm$ n/a & 75.05\% & 82.84\% & 88.53\% & \textbf{95.05\%}\\
& \netdirallfit & \textbf{1.38} $\pm$ 1.33 & 75.11\% & 82.53\% & 88.26\% & 94.58\%\\
\hline
\multirow{3}{*}{\rotatebox[origin=c]{90}{\parbox{1.8em}{\textit{CV\\jun.}}}}
& Lindner et al. (2015) & 1.20 $\pm$ n/a & 84.70\% & 89.38\% & 92.62\% & 96.30\% \\
& Zhong et al. (2019) & 1.22 $\pm$ 2.45 & 86.06\% & 90.84\% & 94.04\% & 97.28\%\\
& \netdirallfit & \textbf{0.99} $\pm$ 1.07 & \textbf{89.76}\% & \textbf{93.74\%} & \textbf{95.83\%} & \textbf{97.82}\%\\
\hline
\end{tabular}
\end{center}
\label{tbl:results_skull}
\end{table}

In Tables~\ref{tbl:results_hand} and \ref{tbl:results_skull}, landmark prediction accuracy measured in point error (PE) and robustness measured in the success detection rate per radius ($\text{SDR}_r$) of our method is compared to the state-of-the-art on both the hand and cephalogram dataset, respectively.
The reported values for the state-of-the-art methods are taken from the respective publications~\citep{Ibragimov2014,Lindner2015,Urschler2018,Payer2019,Zhong2019,Chen2019}.

On a three-fold cross-validation (CV) of the dataset of left hand radiographs, our proposed method that uses an anisotropic Gaussian function to model both the target heatmap and the distribution of the heatmap prediction (\netdirallfit) outperforms the previously published methods~\citep{Lindner2015,Urschler2018,Payer2019}.
%
Also, our method is compared to the winning methods of the ISBI Cephalometric X-ray Image Analysis Challenges in 2014~\citep{Ibragimov2014} and 2015~\citep{Lindner2015} as well as recent state-of-the-art methods on this dataset~\citep{Zhong2019,Chen2019}.
On both test sets \textit{Test1} and \textit{Test2}, it can be observed that the methods using CNNs achieve similar results, while our method is the most accurate with the smallest~PE.
In terms of SDR, our method performs in-line with other methods, while \cite{Chen2019} perform slightly better in \textit{Test1} and \cite{Zhong2019} in \textit{Test2}, respectively.
Nevertheless, the results on \textit{Test1} and \textit{Test2} should be treated with caution because of the systematic shift of some landmarks (e.g. landmark 8 and 16) as compared to the annotations of the training data. 
The systematic shift has already been reported in ~\citep{Lindner2016,Zhong2019} as well as in our MICCAI-UNSURE 2020 workshop paper~\citep{payer2020uncertainty}.  
To mitigate the systematic shift, we follow previous work~\citep{Lindner2016,Zhong2019} and perform a four-fold cross-validation experiment trained and evaluated on the junior annotations only (\textit{CV jun.}). 
Here, similar to the hand radiographs, it can be observed that our method that uses an anisotropic Gaussian function to model both the target heatmap and the distribution of the heatmap prediction (\netdirallfit) leads to the best results, outperforming all other methods.

\subsection{Correspondence of Heatmap Uncertainty and Localization Error}
\label{sec:heatmap_uncertainty}

\newcommand{\simgapeplotwidth}{0.29\textwidth}

\begin{figure}[t]
\centering
\begin{tabular}{ccc}
\rotatebox[origin=l]{90}{\hspace{0em}\vphantom{p}{\qquad $\targetcovariance$-target}}
\includegraphics[width=\simgapeplotwidth]{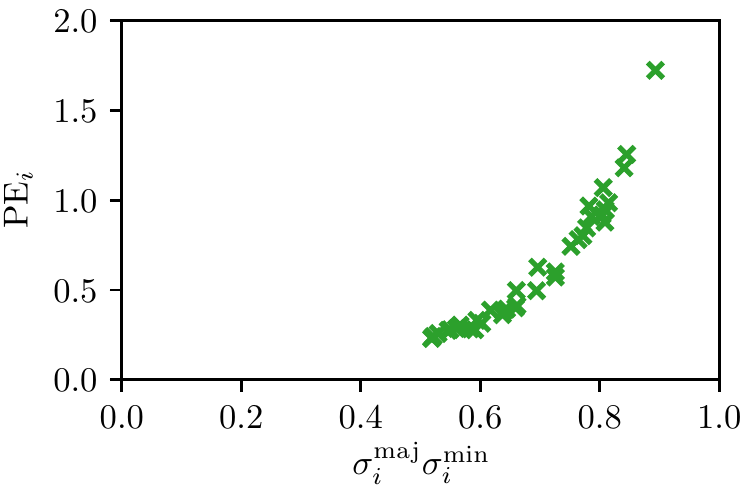} & \includegraphics[width=\simgapeplotwidth]{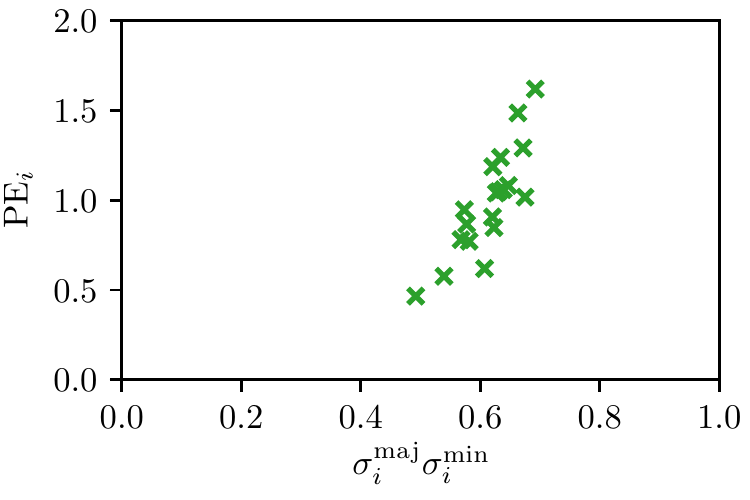} & \includegraphics[width=\simgapeplotwidth]{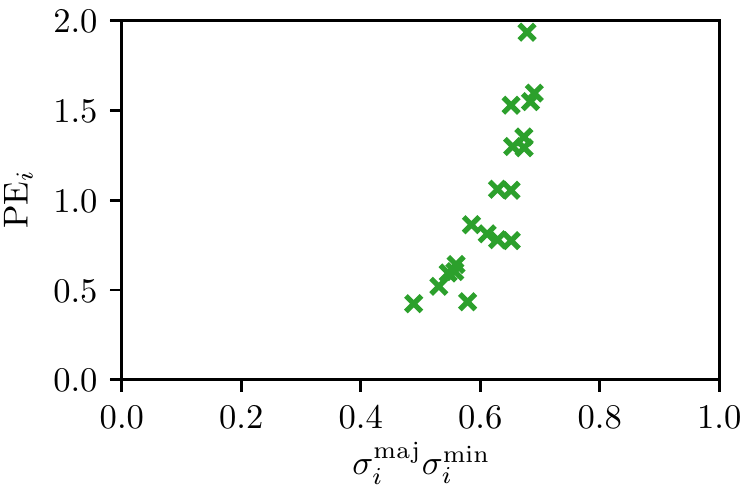}\\
\rotatebox[origin=l]{90}{\hspace{0em}\vphantom{p}{\qquad\quad $\predictioncovariance$-fit}}
\includegraphics[width=\simgapeplotwidth]{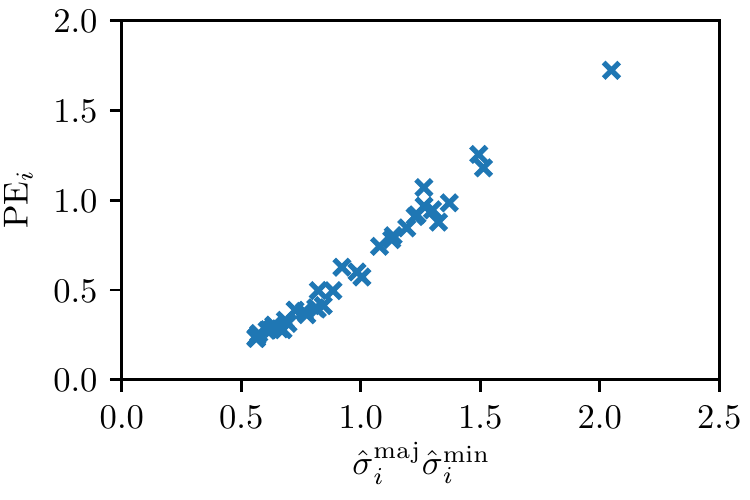} & \includegraphics[width=\simgapeplotwidth]{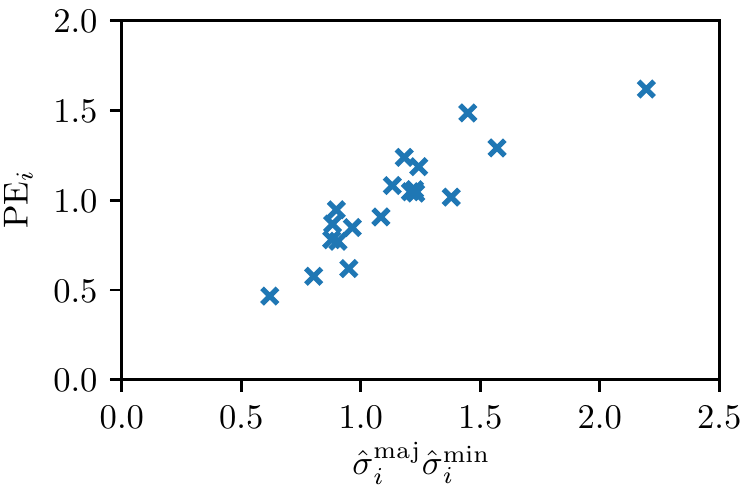} & \includegraphics[width=\simgapeplotwidth]{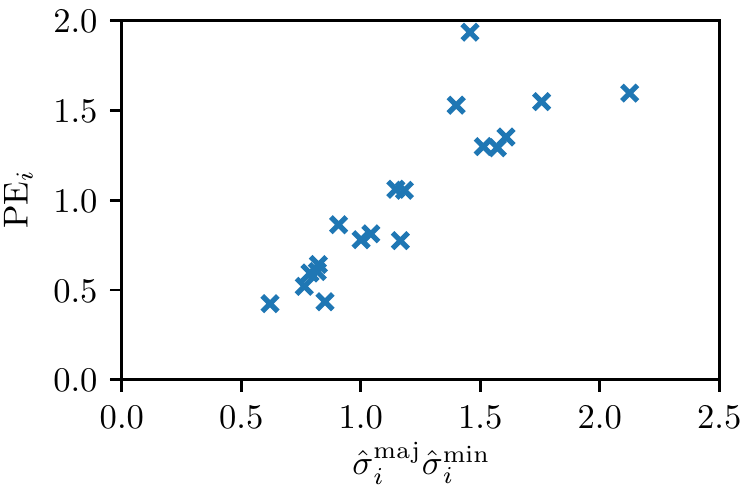}\\
\textit{hand} & \textit{ceph. junior} & \textit{ceph. mean} \\
\end{tabular}
\caption{
Correlation of the mean learned target Gaussian sizes (top) and fitted Gaussian sizes (bottom) of all landmarks $L_i$ to the mean localization error $\text{PE}_i$ for all images of the cross-validation of the hand dataset, as well as the cephalograms with junior annotations (center), and mean of senior and junior (right) in~mm.
The Gaussian sizes are computed as the product $\sigmamajor_i \cdot \sigmaminor_i$, for better visualization, the x-axis of the top and bottom row are not aligned.
}
\label{fig:results:sigma_error_plot}
\end{figure}

\newcommand{\heatmapplotwidth}{0.45\textwidth}
\newcommand{\fiterrorplotwidth}{0.45\textwidth}

\begin{figure}[t]
\centering
\begin{tabular}{cccccc}
\hspace{1.3em} Input
& \hspace{2.0em} $\predictioncovariance$-fit
& \hspace{0.7em} $\targetcovariance$-target
& \hspace{0.8em} Offsets
& \hspace{2.2em} $\predictioncovariance$-fit
& \hspace{0.4em} $\targetcovariance$-target \\
\multicolumn{3}{c}{\includegraphics[width=\heatmapplotwidth]{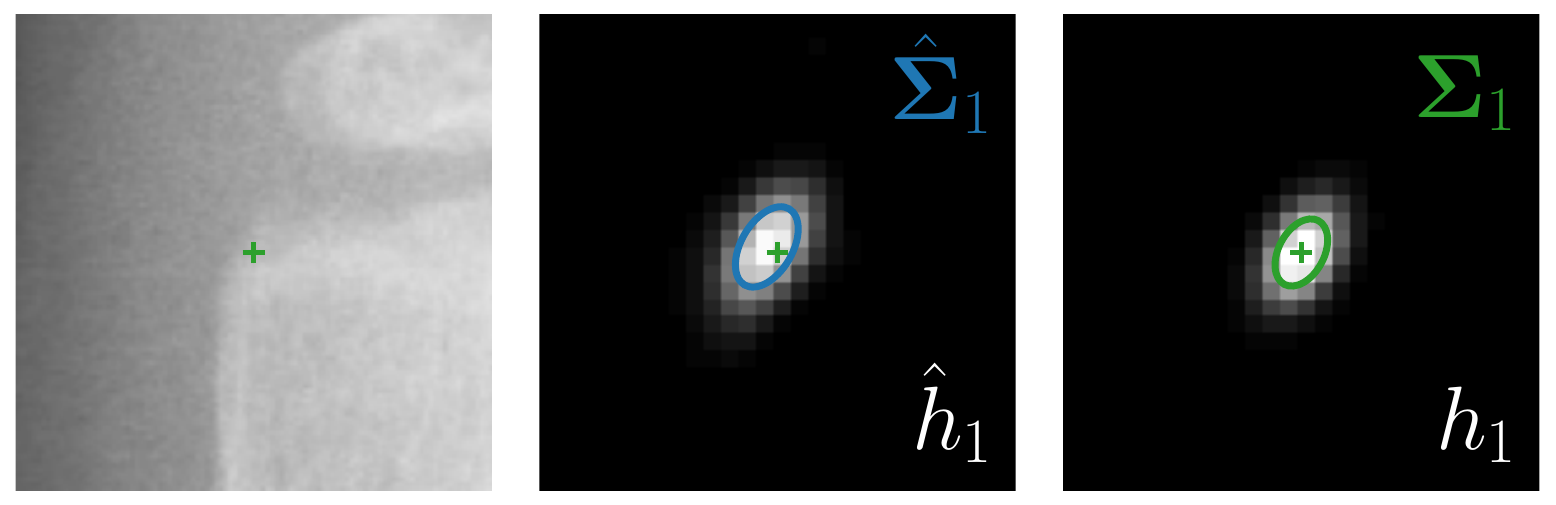}}
& \multicolumn{3}{c}{\includegraphics[width=\fiterrorplotwidth]{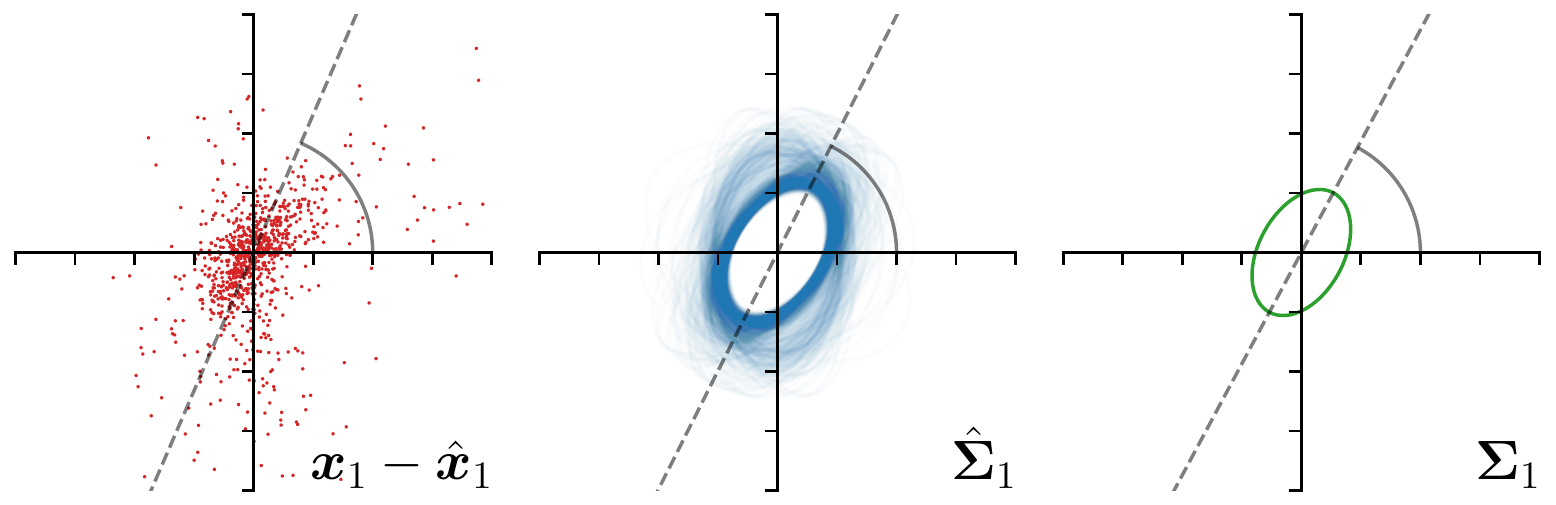}} \\ 
\multicolumn{3}{c}{\includegraphics[width=\heatmapplotwidth]{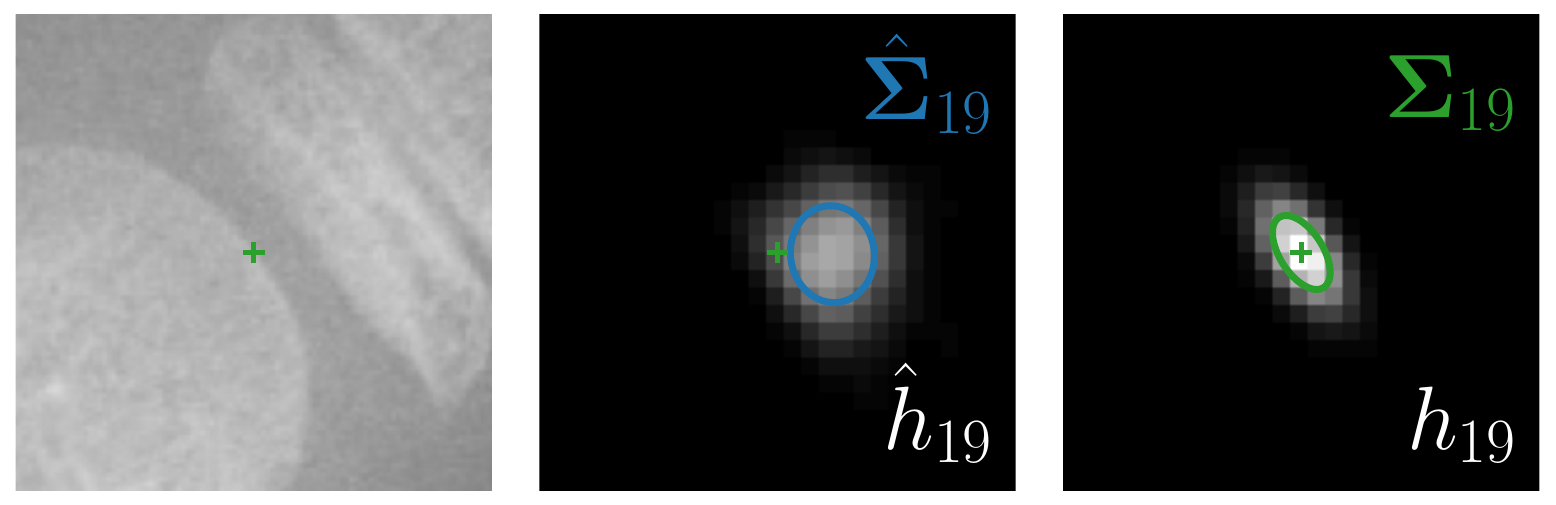}}
& \multicolumn{3}{c}{\includegraphics[width=\fiterrorplotwidth]{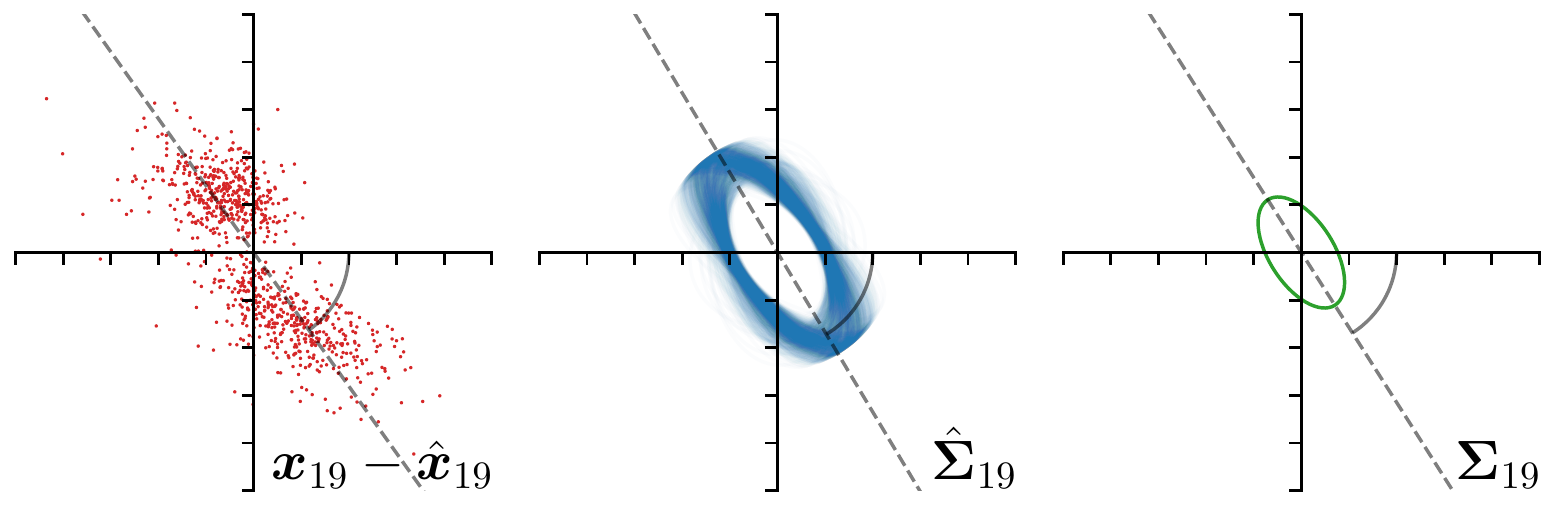}} \\
\multicolumn{3}{c}{\includegraphics[width=\heatmapplotwidth]{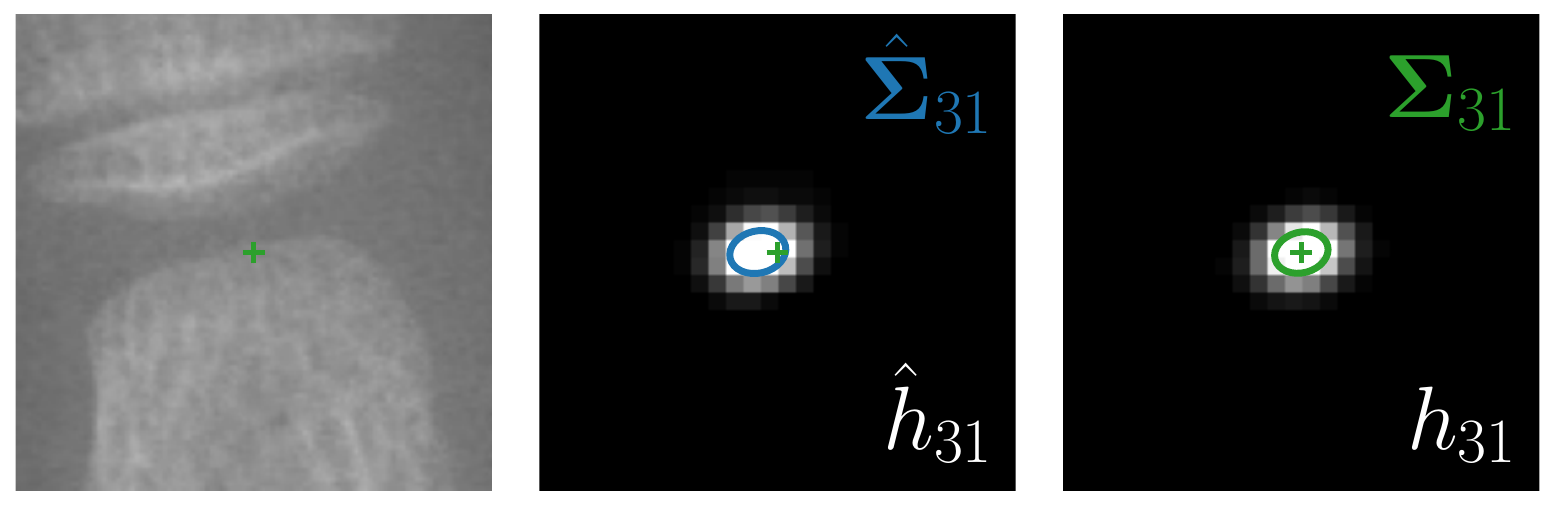}}
& \multicolumn{3}{c}{\includegraphics[width=\fiterrorplotwidth]{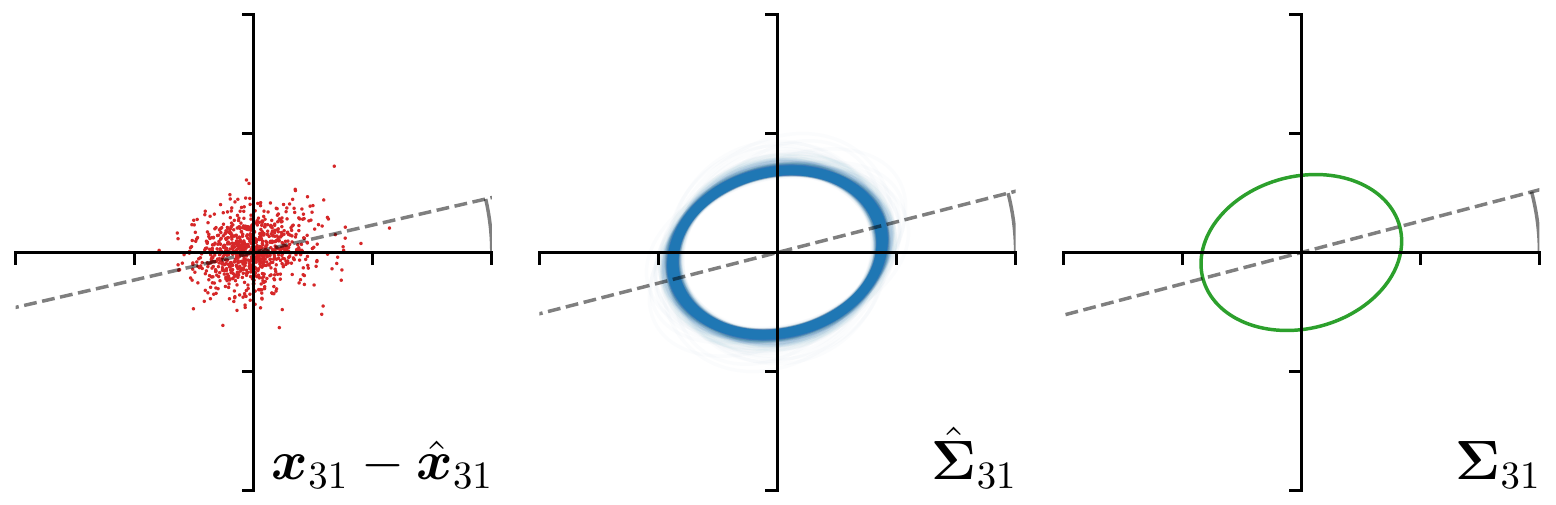}} \\
\multicolumn{3}{c}{\includegraphics[width=\heatmapplotwidth]{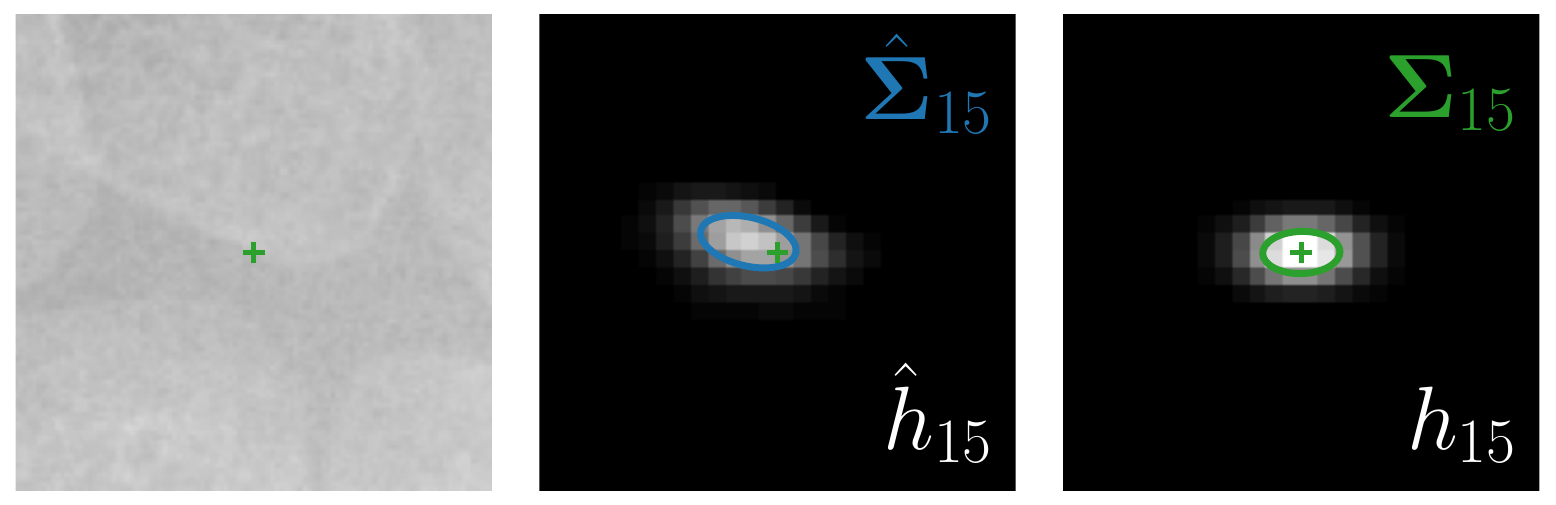}}
& \multicolumn{3}{c}{\includegraphics[width=\fiterrorplotwidth]{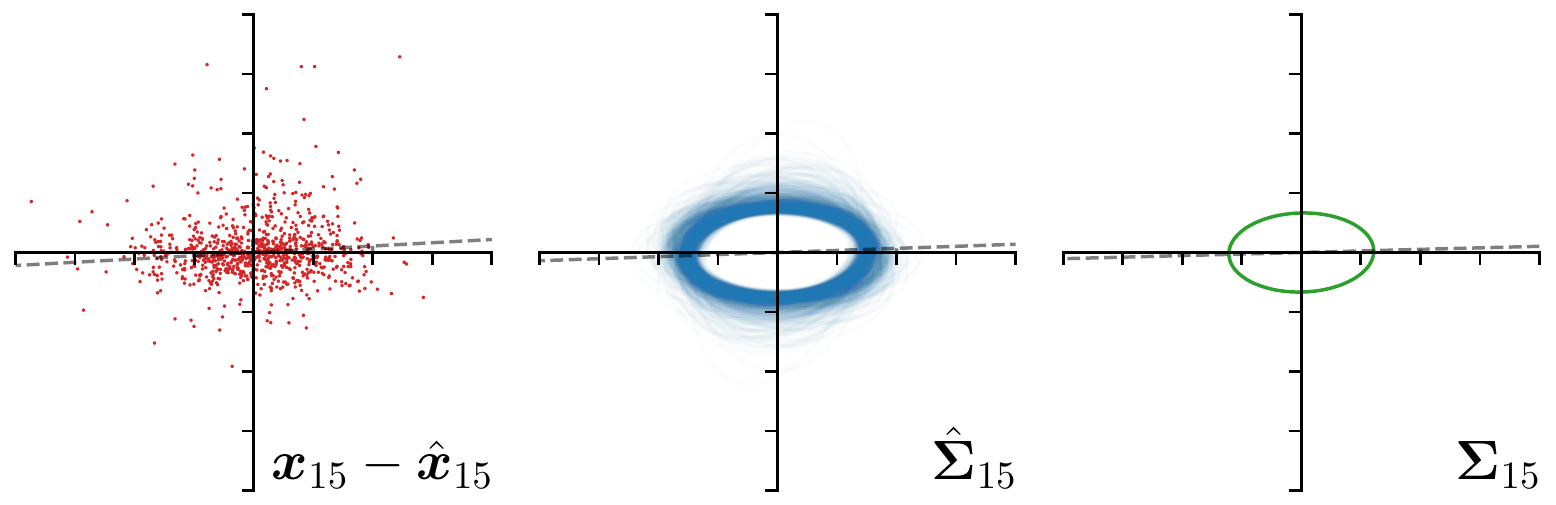}} \\
\end{tabular}
\caption{
Prediction examples for representative landmarks of the dataset of hand X-rays.
Input images, target and predicted heatmaps for an example image are shown in the first three columns.
The groundtruth-prediction offsets for all images are visualized as red dots, while the orientation of a Gaussian fitted to these offsets is shown in gray.
In column $\predictioncovariance$-fit, the distributions of all images fitted to the predicted heatmaps are shown as blue ellipses.
In column $\targetcovariance$-target, the final target heatmaps are shown as green ellipses.
The average rotation of the ellipse's major axis is shown in gray.
}
\label{fig:results:heatmap_plot_hand}
\end{figure}

\begin{figure}[t]
\centering
\begin{tabular}{cccccc}
\hspace{1.3em} Input
& \hspace{2.0em} $\predictioncovariance$-fit
& \hspace{0.7em} $\targetcovariance$-target
& \hspace{0.8em} Offsets
& \hspace{2.2em} $\predictioncovariance$-fit
& \hspace{0.4em} $\targetcovariance$-target \\
\multicolumn{3}{c}{\includegraphics[width=\heatmapplotwidth]{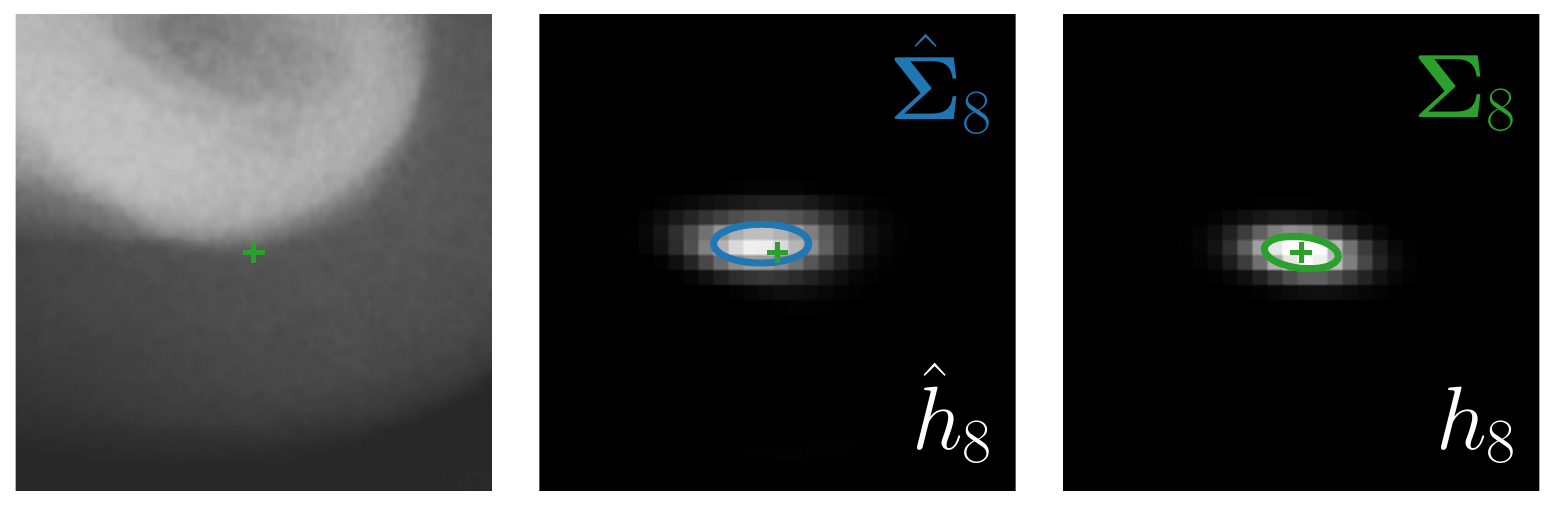}}
& \multicolumn{3}{c}{\includegraphics[width=\fiterrorplotwidth]{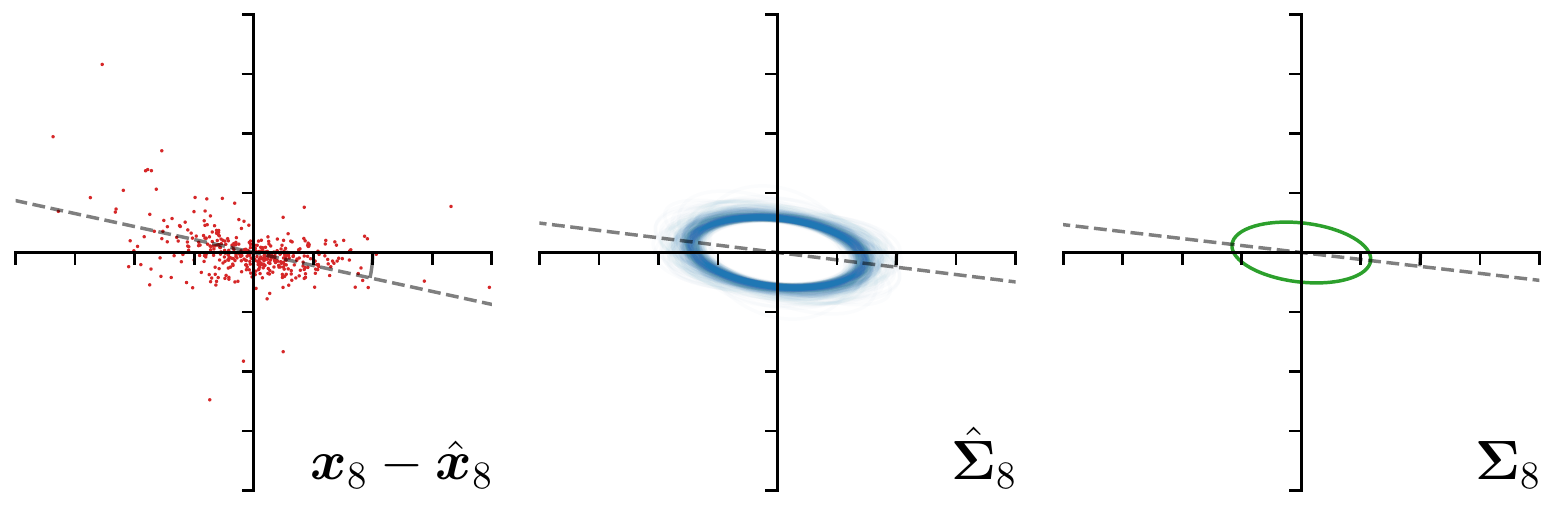}} \\
\multicolumn{3}{c}{\includegraphics[width=\heatmapplotwidth]{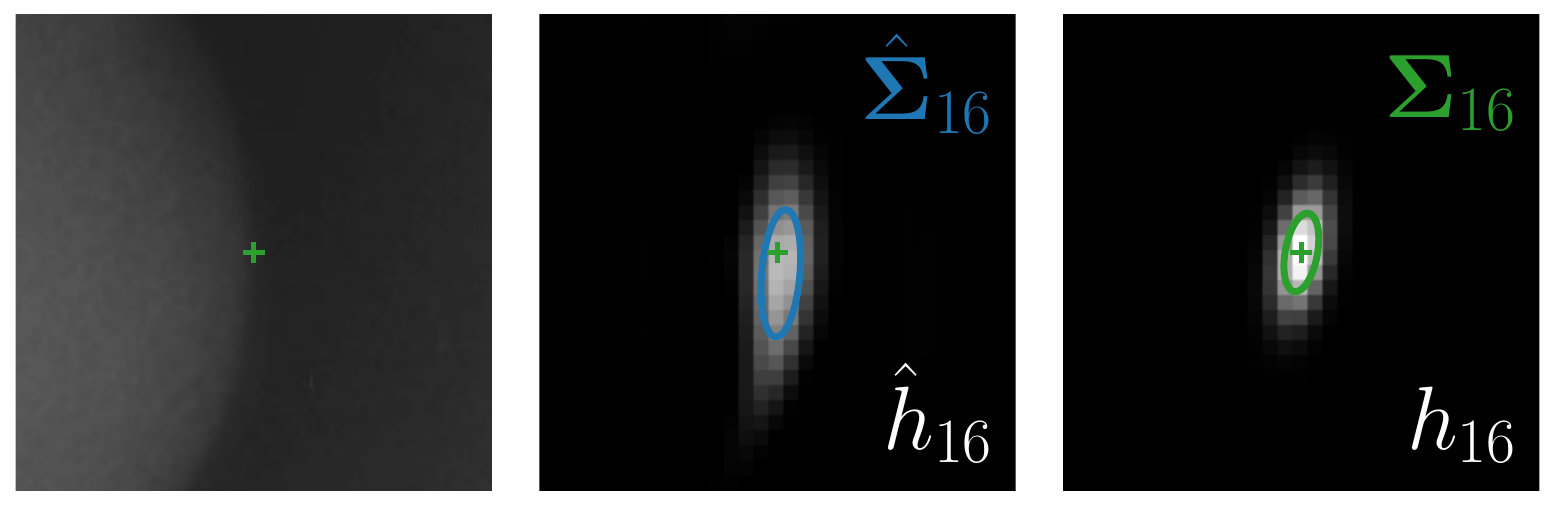}}
& \multicolumn{3}{c}{\includegraphics[width=\fiterrorplotwidth]{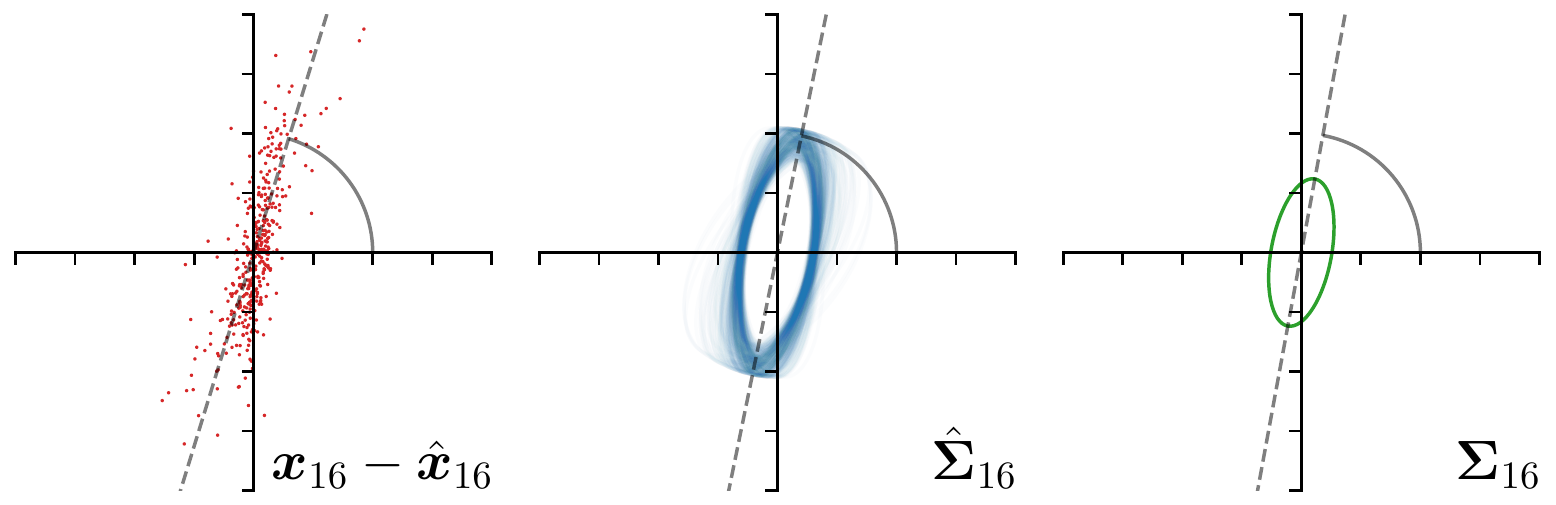}} \\
\multicolumn{3}{c}{\includegraphics[width=\heatmapplotwidth]{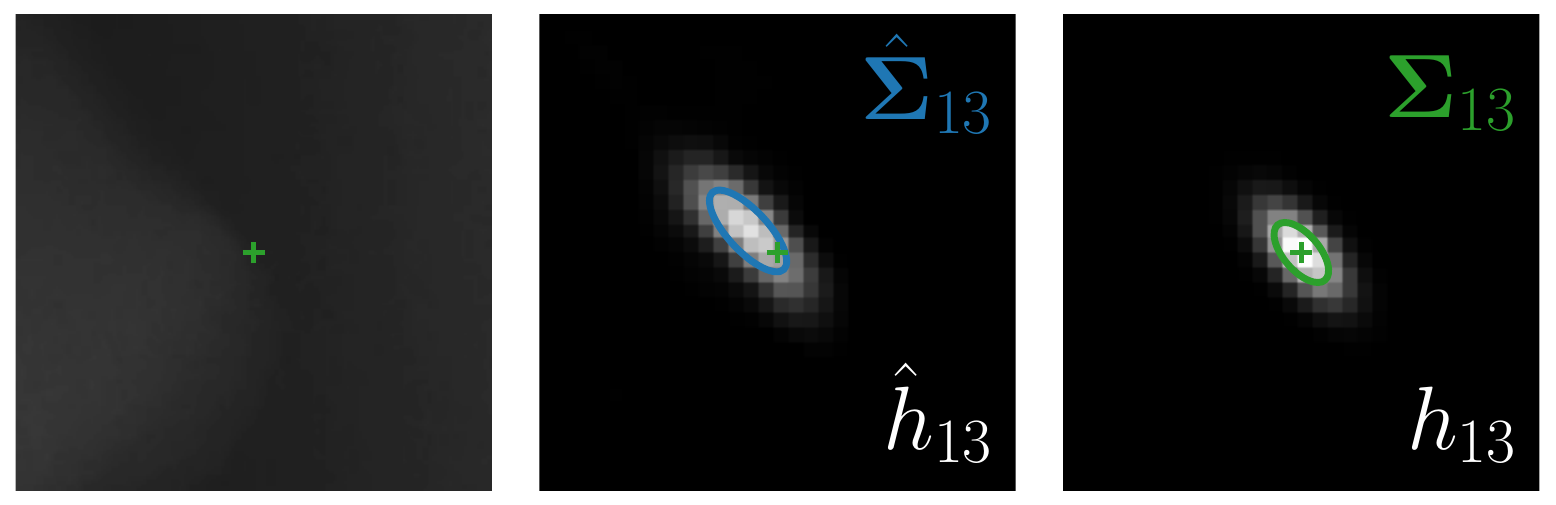}}
& \multicolumn{3}{c}{\includegraphics[width=\fiterrorplotwidth]{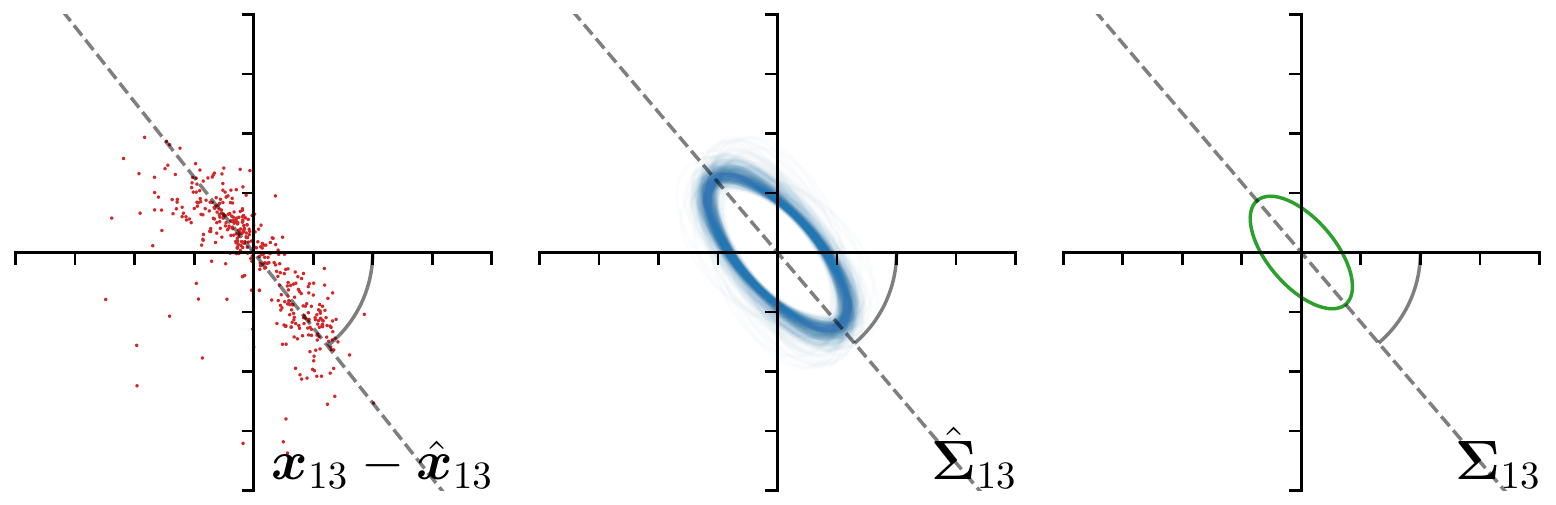}} \\
\multicolumn{3}{c}{\includegraphics[width=\heatmapplotwidth]{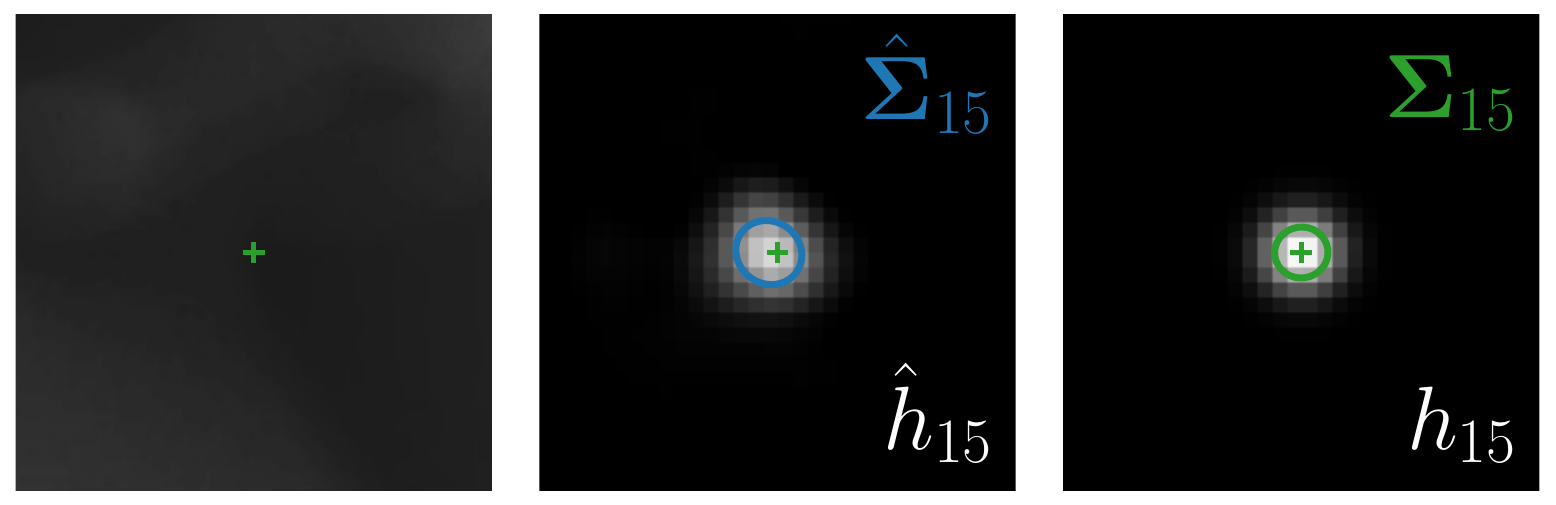}}
& \multicolumn{3}{c}{\includegraphics[width=\fiterrorplotwidth]{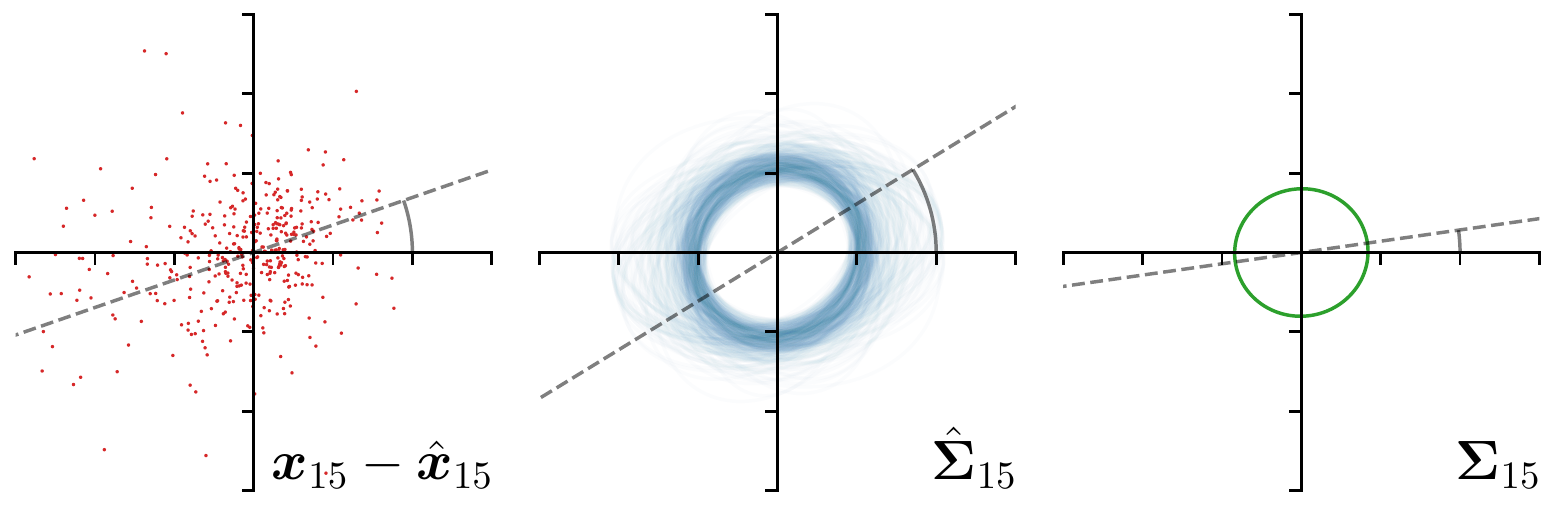}} \\
\end{tabular}
\caption{
Prediction examples for representative landmarks of the dataset of lateral cephalograms.
The images and plots are visualized the same as described in Fig.~\ref{fig:results:heatmap_plot_hand}.
}
\label{fig:results:heatmap_plot_skull}
\end{figure}

In this work, we proposed to use parameters of the target Gaussian heatmaps to model a \textit{dataset-based annotation uncertainty} (see Sec.~\ref{sec:heatmap_regression}) and parameters of a fitted Gaussian function to the predicted heatmaps to model a \textit{sample-based annotation uncertainty} (see Sec.~\ref{sec:heatmap_fitting}). 
In this section, it is shown that both sets of Gaussian parameters represent valid uncertainty measures by comparing the sizes of the Gaussians with the corresponding localization error for each landmark individually (see Fig.~\ref{fig:results:sigma_error_plot}).
For both the hand and cephalogram dataset, the size of the learned target Gaussians as well as the size of the fitted Gaussians are calculated as the product of the variances ($\sigmamajor \cdot \sigmaminor$) averaged over the cross-validations or over all test images, respectively.
It can be observed that both the target and fitted heatmap parameters correlate with the PE. 

Our proposed method is also able to model directional uncertainties, 
which is visualized in Fig.~\ref{fig:results:heatmap_plot_hand} and~\ref{fig:results:heatmap_plot_skull} for selected landmarks. 
On the left side of the figures, the target and predicted heatmaps are superimposed with ellipses representing the target and fitted Gaussian parameters, respectively.
Both target and predicted heatmaps are smaller for landmarks on distinct corners as compared to landmarks on edges.
For landmarks on edges it can be observed that the heatmaps are oriented in direction of the underlying edges.
Thus, for these landmarks the main sources for ambiguity are along the edges.
In Fig.~\ref{fig:results:heatmap_plot_hand} and~\ref{fig:results:heatmap_plot_skull} (right) it is shown that both the learned target heatmap parameters $\targetcovariance_i$ and the fitted heatmap parameters $\predictioncovariance_i$ correspond in orientation and size to the expected prediction-groundtruth offsets $\targetcoordinate_i - \predictioncoordinate_i$.
For each landmark individually, the parameters of both target and predicted heatmaps are represented as ellipses centered at the origin, and plotted together for all images.
To visualize the prediction-groundtruth offsets, individual predictions for all images are shown as points relative to their respective groundtruth positions, centered at the origin.
From Fig.~\ref{fig:results:heatmap_plot_hand} and~\ref{fig:results:heatmap_plot_skull} it can be observed that the orientation and the size of both predicted and target heatmap parameters correspond to the prediction-groundtruth offsets.
Overall, these results confirm that the target and the fitted parameters of the Gaussian functions are valid uncertainty measures.

\subsection{Correspondence of Heatmap Uncertainty and Annotation Distribution}
\label{sec:inter}

\newcommand{\smallinterplotwidth}{0.19\textwidth}
\newcolumntype{?}[1]{!{\vrule width #1}}

\begin{figure}[t]
\centering
\begin{tabular}{ccccc}
& Ann. & Ann.-fit & $\predictioncovariance$-fit & $\targetcovariance$-target \\

\rotatebox[origin=l]{90}{\hspace{3.5em}\vphantom{p}{$L_1$}}
& \includegraphics[width=\smallinterplotwidth]{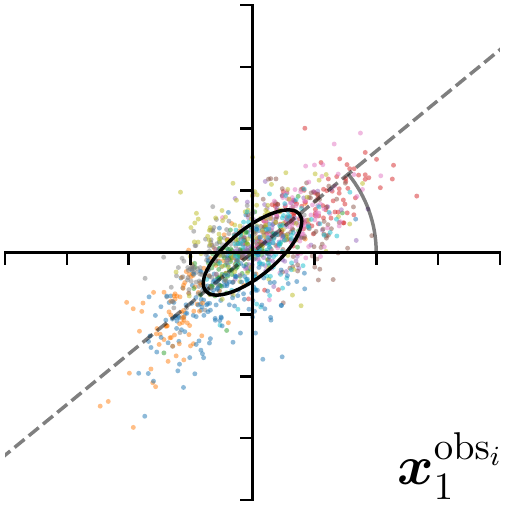} 
& \includegraphics[width=\smallinterplotwidth]{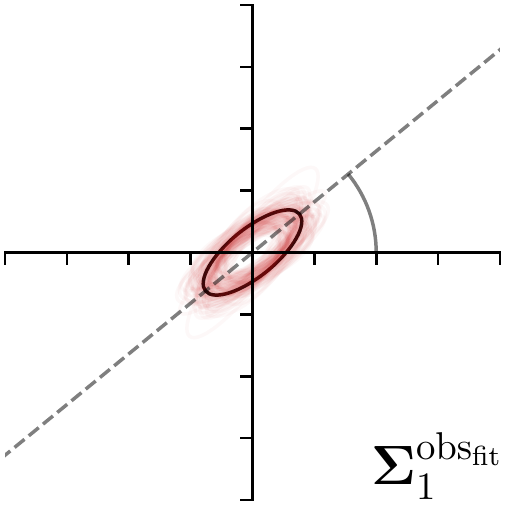}
& \includegraphics[width=\smallinterplotwidth]{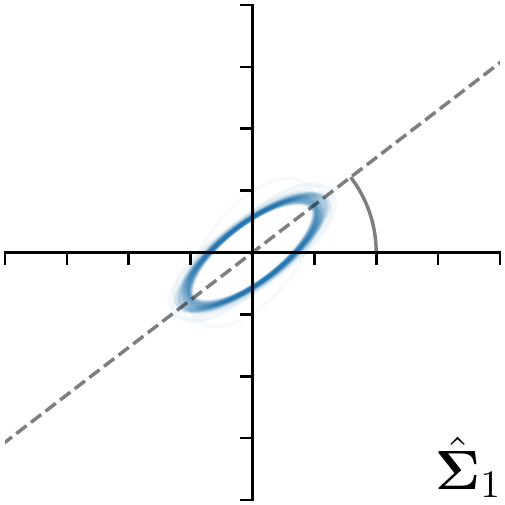} 
& \includegraphics[width=\smallinterplotwidth]{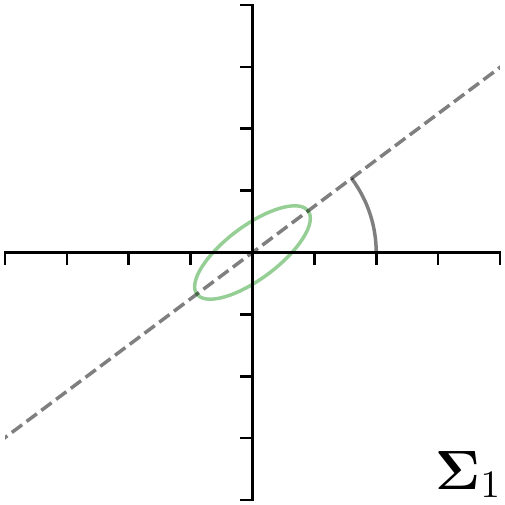} \\

\rotatebox[origin=l]{90}{\hspace{3.5em}\vphantom{p}{$L_2$}}
& \includegraphics[width=\smallinterplotwidth]{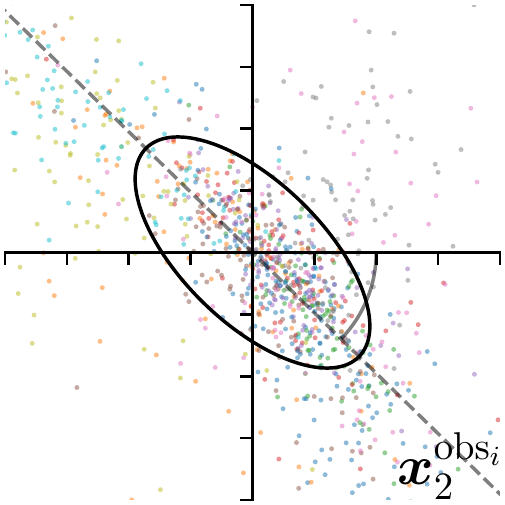} 
& \includegraphics[width=\smallinterplotwidth]{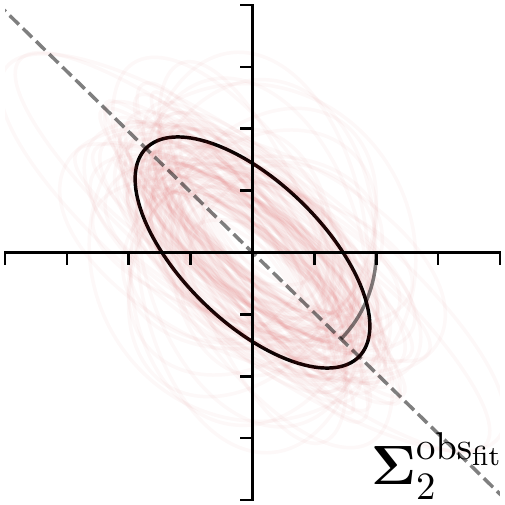}
& \includegraphics[width=\smallinterplotwidth]{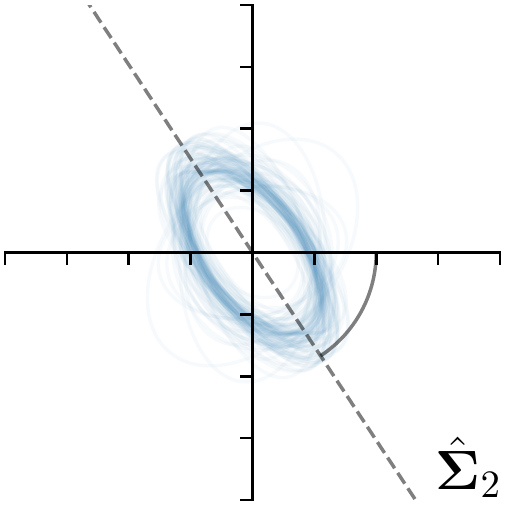} 
& \includegraphics[width=\smallinterplotwidth]{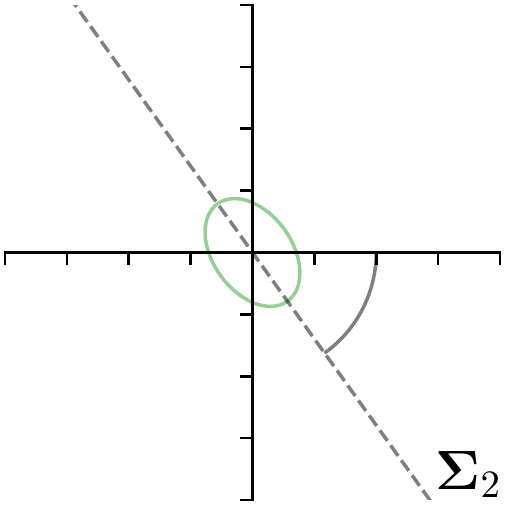} \\

\rotatebox[origin=l]{90}{\hspace{3.5em}\vphantom{p}{$L_3$}}
& \includegraphics[width=\smallinterplotwidth]{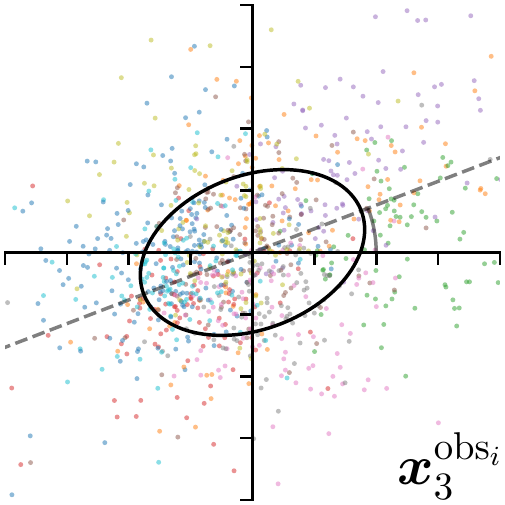} 
& \includegraphics[width=\smallinterplotwidth]{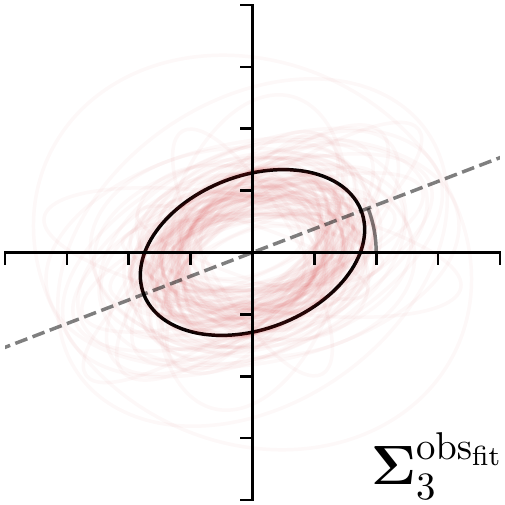}
& \includegraphics[width=\smallinterplotwidth]{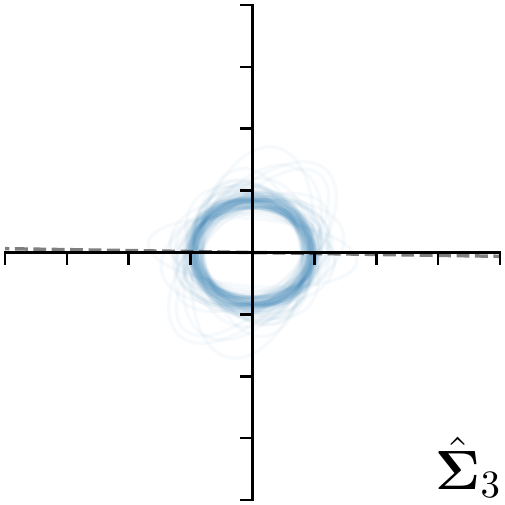} 
& \includegraphics[width=\smallinterplotwidth]{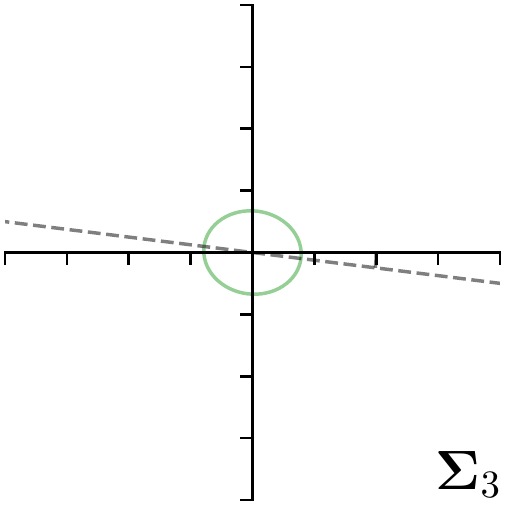} \\

\rotatebox[origin=l]{90}{\hspace{3.5em}\vphantom{p}{$L_4$}}
& \includegraphics[width=\smallinterplotwidth]{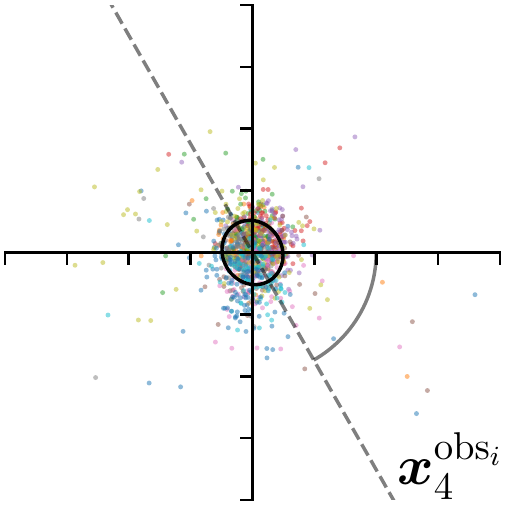} 
& \includegraphics[width=\smallinterplotwidth]{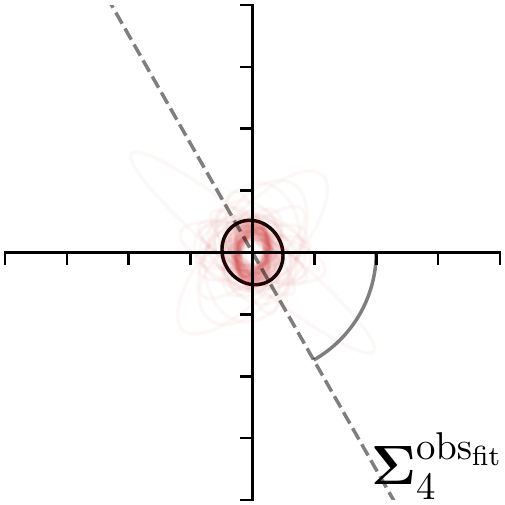}
& \includegraphics[width=\smallinterplotwidth]{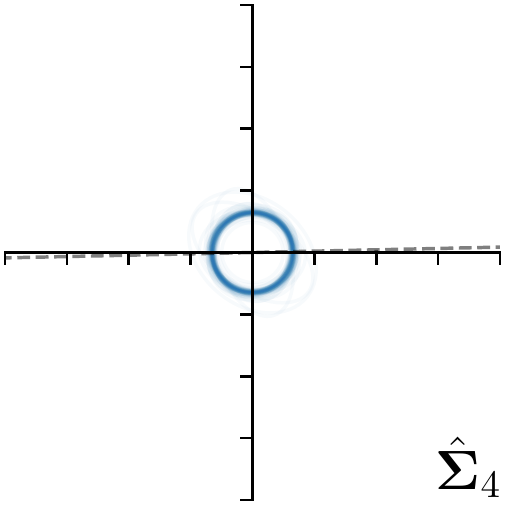} 
& \includegraphics[width=\smallinterplotwidth]{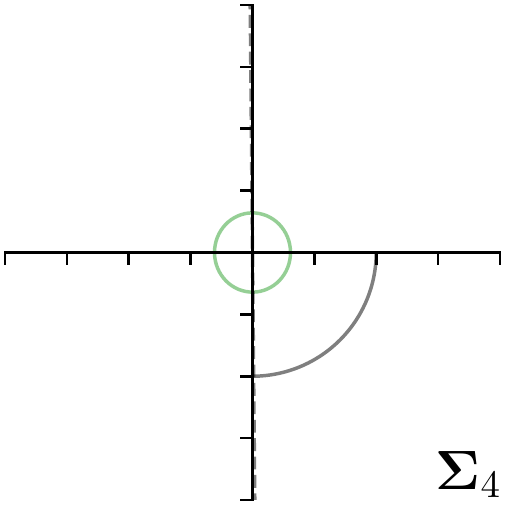} \\

\rotatebox[origin=l]{90}{\hspace{3.5em}\vphantom{p}{$L_5$}}
& \includegraphics[width=\smallinterplotwidth]{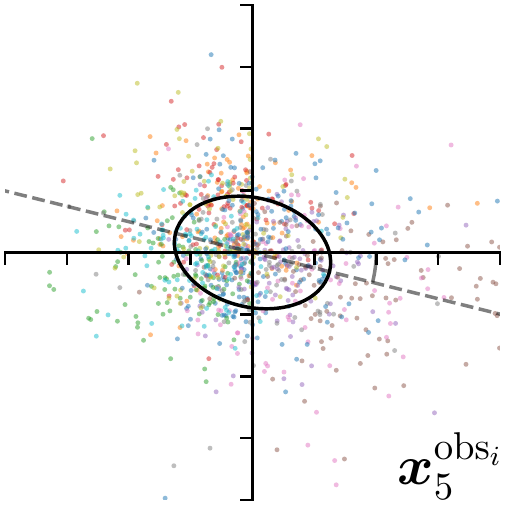} 
& \includegraphics[width=\smallinterplotwidth]{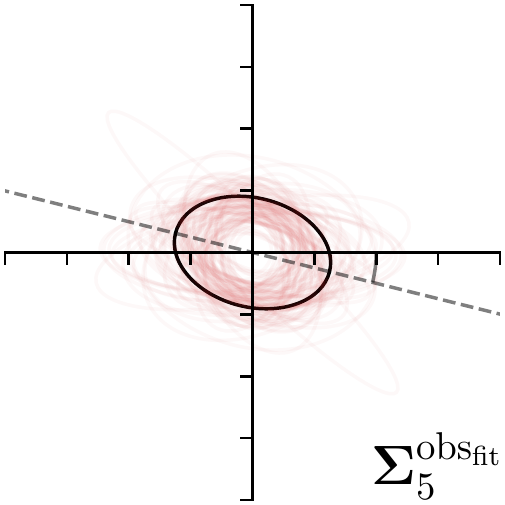}
& \includegraphics[width=\smallinterplotwidth]{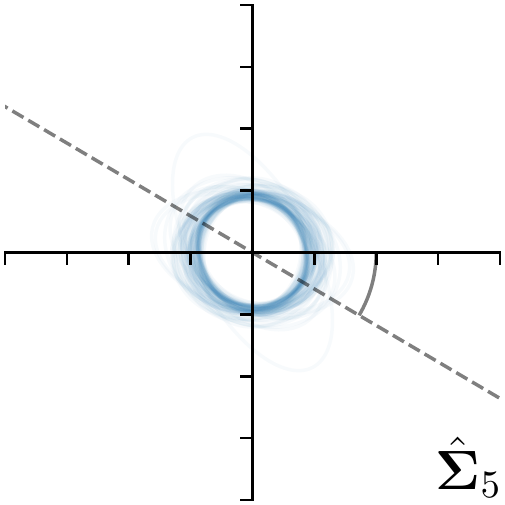} 
& \includegraphics[width=\smallinterplotwidth]{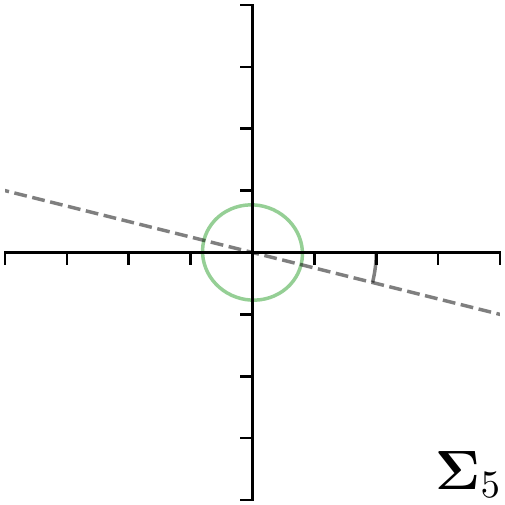} \\
\end{tabular}
\caption{
Annotation distributions as well as predicted distributions from our method for the five landmarks with 11 annotations.
In column Ann., the individual offsets for all 11 annotators to the mean for each image are shown as colored dots, while Ann.-fit shows the individual fitted distributions for all 100 images as red ellipses.
The average distribution is shown as a black ellipse.
$\predictioncovariance$-fit and $\targetcovariance$-target are visualized the same as in Fig.~\ref{fig:results:heatmap_plot_hand}.
}
\label{fig:inter_landmarks}
\end{figure}

To reliably assess inter-observer variability, neither the hand dataset nor the cephalogram dataset can be used, as they were annotated by only one or two annotators per image, respectively.
Therefore, we randomly chose 100 images of the cephalogram dataset and employed nine annotators experienced in medical image analysis to label five characteristic landmarks on each image.
Combined with the junior and senior radiologist annotations already existing in the cephalogram dataset, this leads to a total of 11 annotations per image and landmark.

We trained a model of our proposed method ($\anisoaniso$) on the mean coordinate according to the 11 annotations per image.
The annotation and predicted distributions of the selected points are compared in Fig.~\ref{fig:inter_landmarks}.
Similarly to how the prediction-groundtruth offsets are presented in Fig.~\ref{fig:results:heatmap_plot_hand}, the annotation offsets for each annotator and image are visualized as points relative to the respective mean annotation of all 11 annotators, centered at the origin (Fig.~\ref{fig:inter_landmarks}, first column).
To model the annotation distribution, a Gaussian distribution is fitted to the 11 annotations for each landmark and image individually.
All annotation distributions as well as their mean are visualized as ellipses (Fig.~\ref{fig:inter_landmarks}, second column).
The parameters of both predicted heatmaps (Fig.~\ref{fig:inter_landmarks}, third column) and target heatmaps (Fig.~\ref{fig:inter_landmarks}, last column) are represented as ellipses centered at the origin, plotted for all images.
By visually comparing the size and the orientation of both predicted and target heatmap parameters with the annotation distributions, a correspondence can clearly be seen.

\begin{table}[t]
\caption{
Gaussian parameters of the annotation distributions (Ann.), fitted to the predicted heatmaps ($\predictioncovariance$-fit), and target heatmaps ($\targetcovariance$-target).
The product $\sigmamajor \cdot \sigmaminor$, the ratio $\sigmamajor : \sigmaminor$ and the rotation angle $\theta$ of the respective distributions are shown.
The mean and standard deviation is shown where applicable.
For landmarks with low anisotropy and less indicative rotation ($L_3$, $L_4$, and $L_5$), the rotation angles are shown in gray.
}
\begin{center}
\scriptsize
\begin{tabular}{ c|c|c|c|c|c|c }
\cline{1-7}
\cline{1-7}
Metric & Method & \multicolumn{1}{c}{$L_1$} & \multicolumn{1}{c}{$L_2$} & \multicolumn{1}{c}{$L_3$} & \multicolumn{1}{c}{$L_4$} & \multicolumn{1}{c}{$L_5$} \\

\hline

\multirow{3}{*}{\parbox{5em}{\centering\mbox{Ratio}\\\mbox{$\sigmamajor:\sigmaminor$}}}
& Ann. & 2.57 & 2.12 & 1.52 & 1.11 & 1.46 \\
& $\predictioncovariance$-fit & 2.80 $\pm$ 0.25 & 1.98 $\pm$ 0.47 & 1.34 $\pm$ 0.30 & 1.12 $\pm$ 0.12 & 1.20 $\pm$ 0.17 \\
& $\targetcovariance$-target & 2.63 & 1.51 & 1.17 & 1.03 & 1.05 \\

\hline

\multirow{3}{*}{\parbox{5em}{\centering\mbox{Product}\\\mbox{$\sigmamajor \cdot \sigmaminor$}}}
& Ann. & 0.37 & 2.73 & 2.34 & 0.26 & 1.13 \\
& $\predictioncovariance$-fit & 0.63 $\pm$ 0.07 & 1.40 $\pm$ 0.35 & 0.86 $\pm$ 0.18 & 0.44 $\pm$ 0.08 & 0.98 $\pm$ 0.24 \\
& $\targetcovariance$-target & 0.48 & 0.62 & 0.52 & 0.40 & 0.62 \\

\hline

\multirow{3}{*}{\parbox{5em}{\centering\mbox{Rotation $\theta$}\\\mbox{(in degree)}}}
& Ann. & 39.33 & -44.30 & \gray{20.97} & \gray{-60.34} & \gray{-13.98} \\
& $\predictioncovariance$-fit & 37.48 $\pm$ 2.72 & -56.62 $\pm$ 14.82 & \gray{-0.88 $\pm$ 35.90} & \gray{1.25 $\pm$ 57.68} & \gray{-30.52 $\pm$ 36.00} \\
& $\targetcovariance$-target & 36.79 & -54.39 & \gray{-7.10} & \gray{-89.47} & \gray{-14.04} \\





\cline{1-7}
\cline{1-7}
\end{tabular}
\label{tb:inter_landmarks}
\end{center}
\end{table}

When comparing the annotation offsets for different landmarks in Fig.~\ref{fig:inter_landmarks}, it can be observed that landmark 4 on the tip of the incisor has the best agreement among the annotators, due to its well defined location (see Fig.~\ref{fig:selected_landmarks}).
Although landmark 5 also lies on an anatomically well defined location, due to low contrast in this region, its annotation variation is higher.
Landmarks 2 and 3 were the most problematic ones to annotate.
Landmark 2 is challenging, because its position has to be estimated by the annotators in between the left and right mandible, when they are not aligned and therefore, are not seen as a single edge in the lateral cephalograms.
Landmark 3 lies on an edge of a bone within the skull that is challenging to recognize and becomes ambiguous if the head is tilted. 
For landmark 1 that lies on the tip of the chin, it can be seen that the annotators correctly identified the edge of the chin, however, identifying the exact position along the edge of the chin was challenging.
Thus, the annotation distribution for this landmark has a clear dominant direction.

In all the previous works based on heatmap regression, the target heatmaps are modeled as isotropic functions.
However, from the annotation distributions in Fig.~\ref{fig:inter_landmarks}, second column, clear anisotropic orientations of the inter-observer variabilities of the landmarks located at smooth edges, i.e., landmark 1 and 2 are observable.
Thus, to estimate the inter-observer variability, in this work, we used anisotropic Gaussian functions for target heatmaps, as well as to model the predicted distribution.
This enables our method to successfully model the inter-observer variability as can be seen when comparing both the predicted Gaussian functions in the third and the learned target Gaussian functions in the last column to the annotation distributions in the second column of Fig.~\ref{fig:inter_landmarks}. 

To quantitatively validate this correspondence between the parameters of the Gaussian functions of our proposed method and the annotation distribution, we present the ratio and product of the extent of the major axis $\sigmamajor$ and minor axis $\sigmaminor$ as well as the orientation angle $\theta$ of the Gaussian functions in Table~\ref{tb:inter_landmarks}.
The product of the extents of the major axis $\sigmamajor$ and minor axis $\sigmaminor$ shows that the parameters of both target and average fitted Gaussians have the tendencies of underestimating the size of the annotation uncertainty for landmarks with a larger inter-observer variability (landmark 2, 3, 5).
Nevertheless, for the landmarks with lower inter-observer variability (landmark 1, 4), the size of the Gaussian functions is in range with the size of the annotation uncertainty.
From the ratio of the extent of the major axis $\sigmamajor$ and minor axis $\sigmaminor$, it can be observed that our method is not only able to model isotropic annotation distributions (landmark 3, 4, 5) but also anisotropic ones (landmark 1, 2).
It is interesting to see that both ratios of the target and the average fitted Gaussian parameters are similar to the ratio of the annotation distributions.
Furthermore, the rotation angles $\theta$ of the Gaussian functions also approximate the rotation angle of the annotation distribution for the anisotropic landmarks well.

\newcommand{\resultswithcovarianceplotwidth}{0.30\textwidth}

\begin{figure}[t]
\centering
\begin{tabular}{cccc}
\includegraphics[width=\resultswithcovarianceplotwidth]{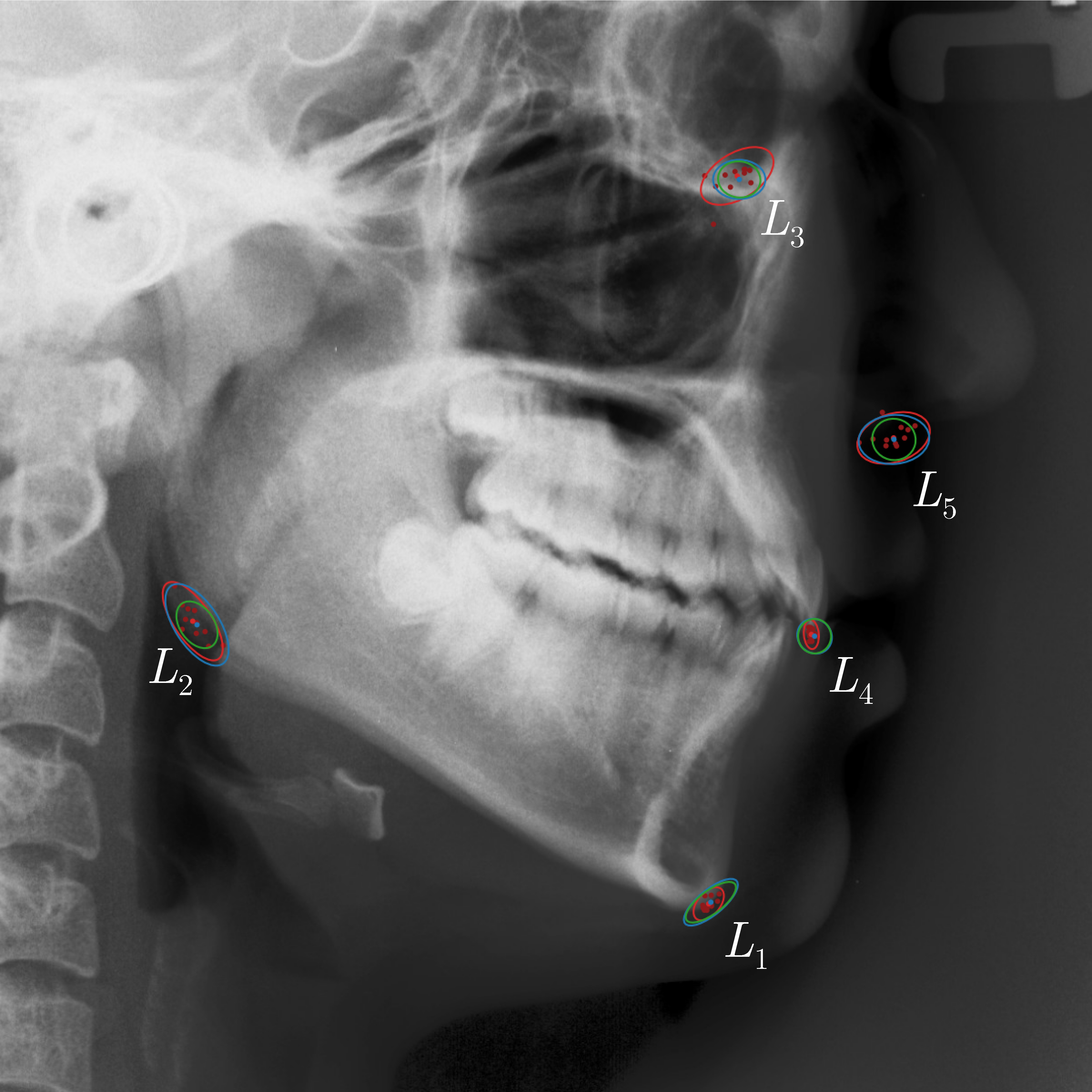} & 
\includegraphics[width=\resultswithcovarianceplotwidth]{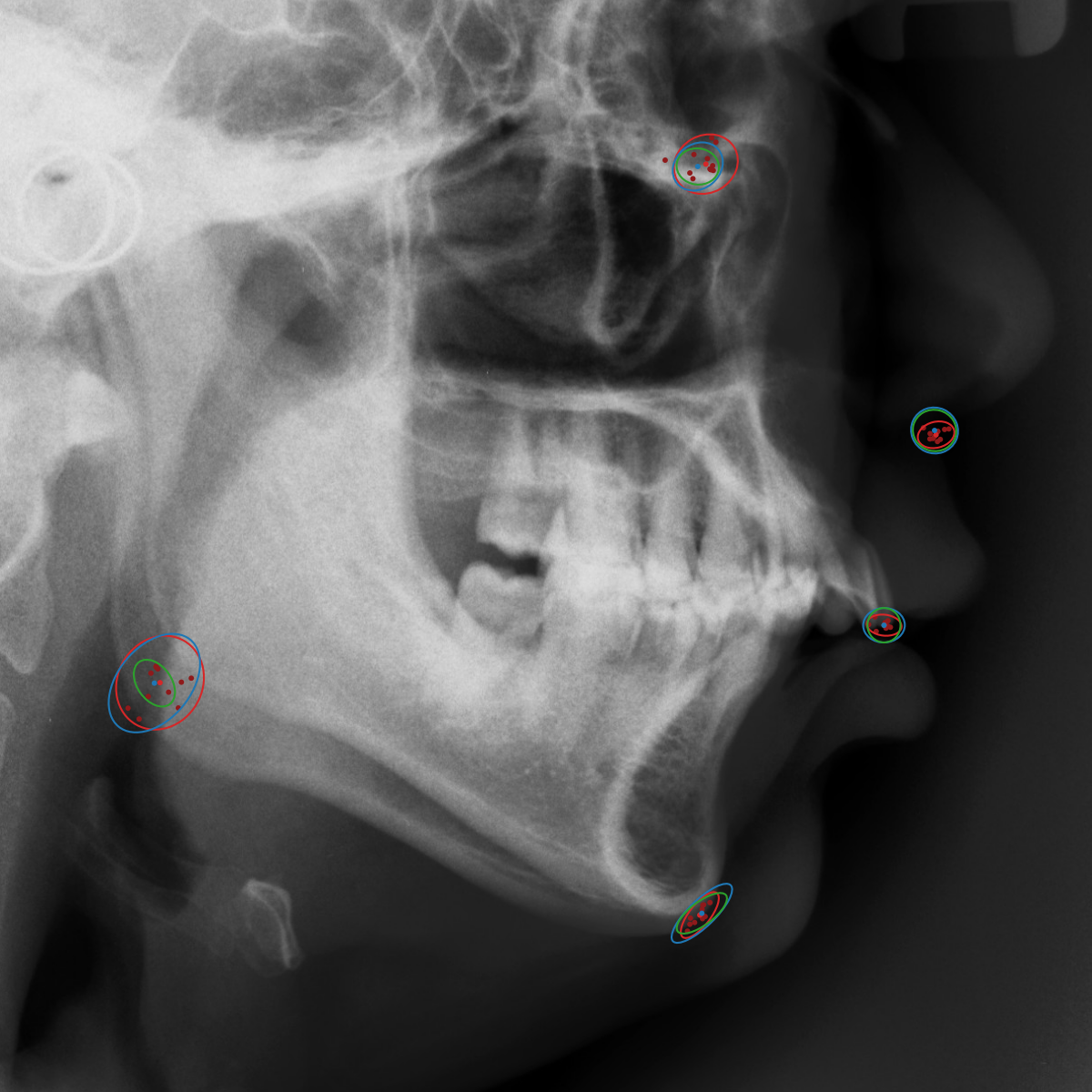} & \includegraphics[width=\resultswithcovarianceplotwidth]{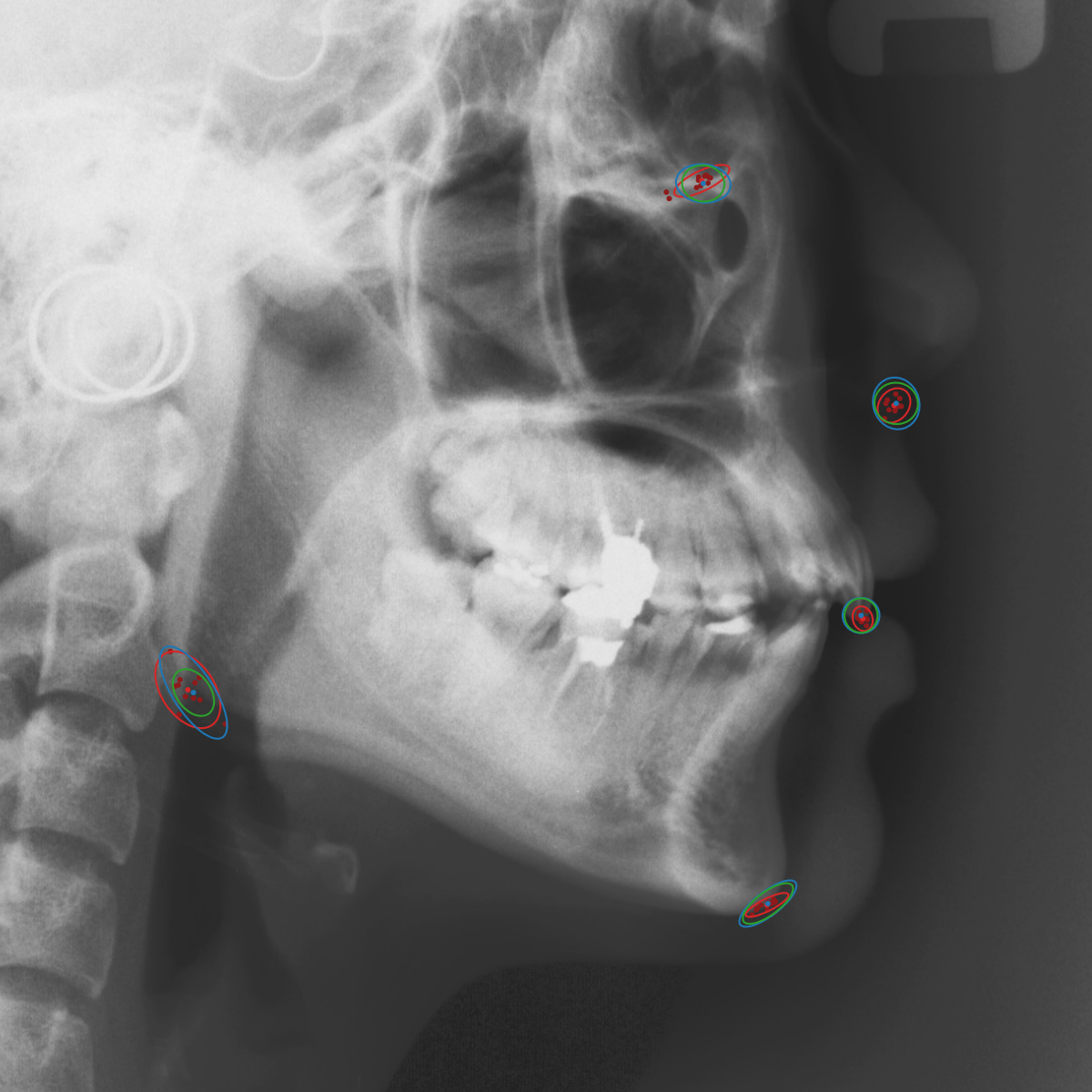} \\
\includegraphics[width=\resultswithcovarianceplotwidth]{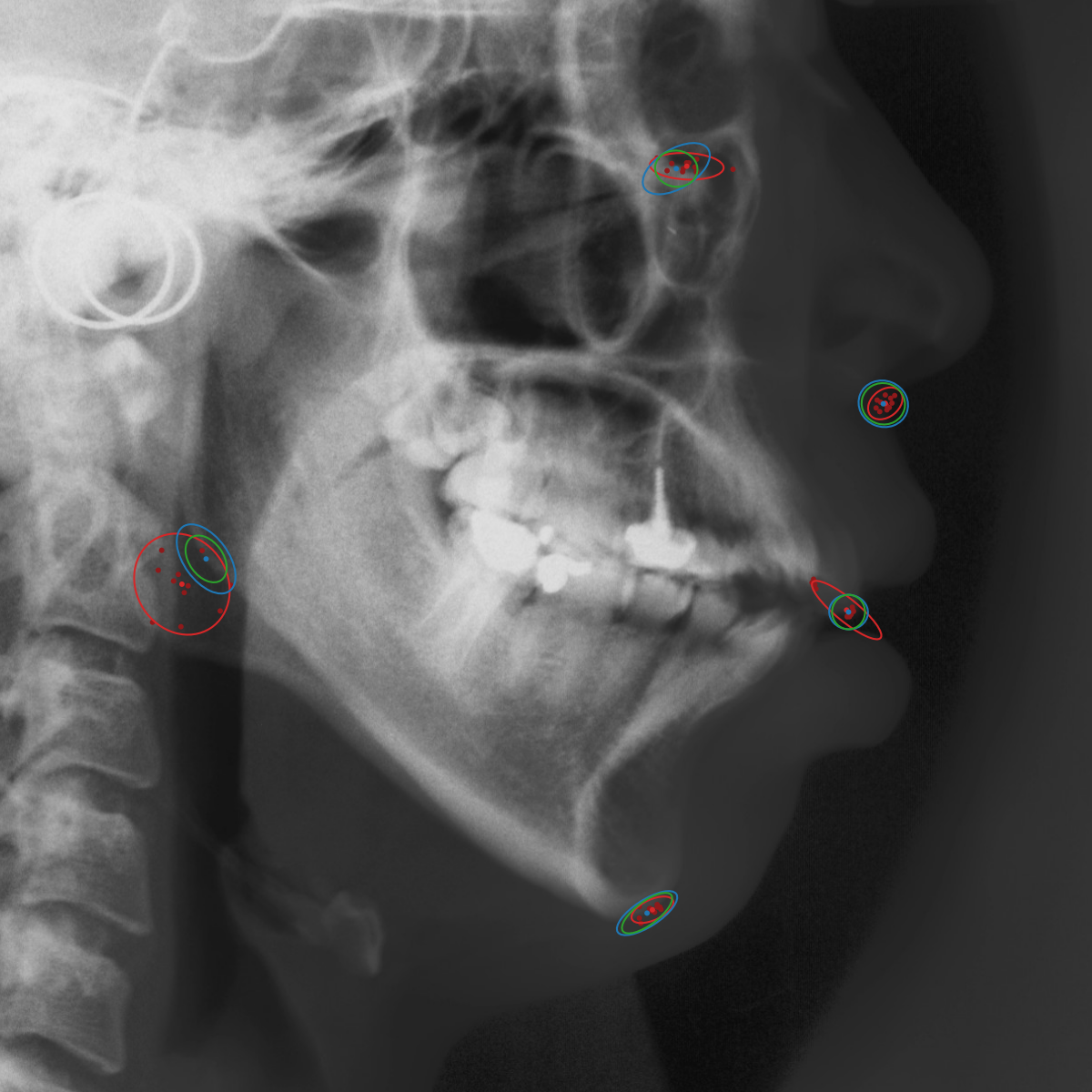} &
\includegraphics[width=\resultswithcovarianceplotwidth]{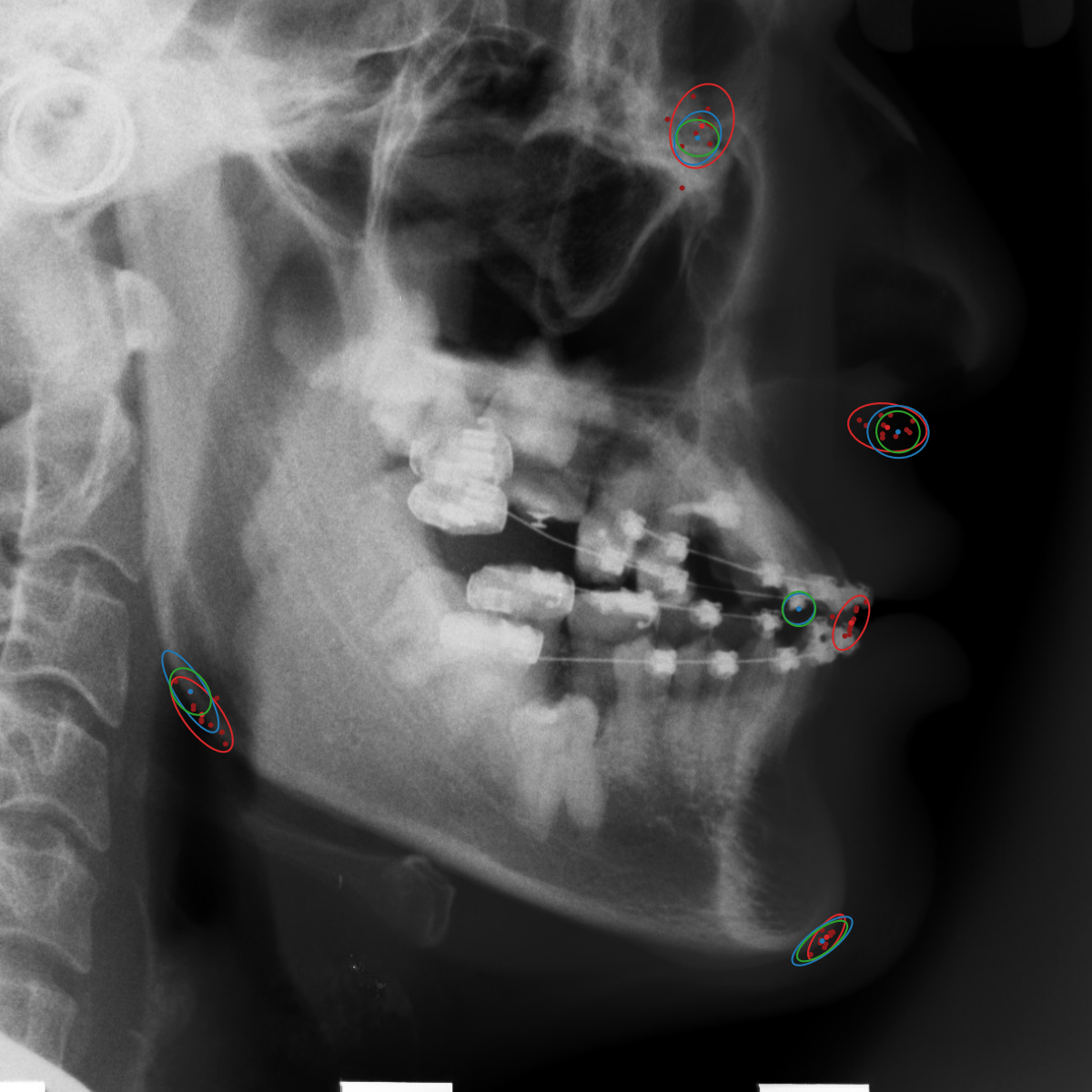} & \includegraphics[width=\resultswithcovarianceplotwidth]{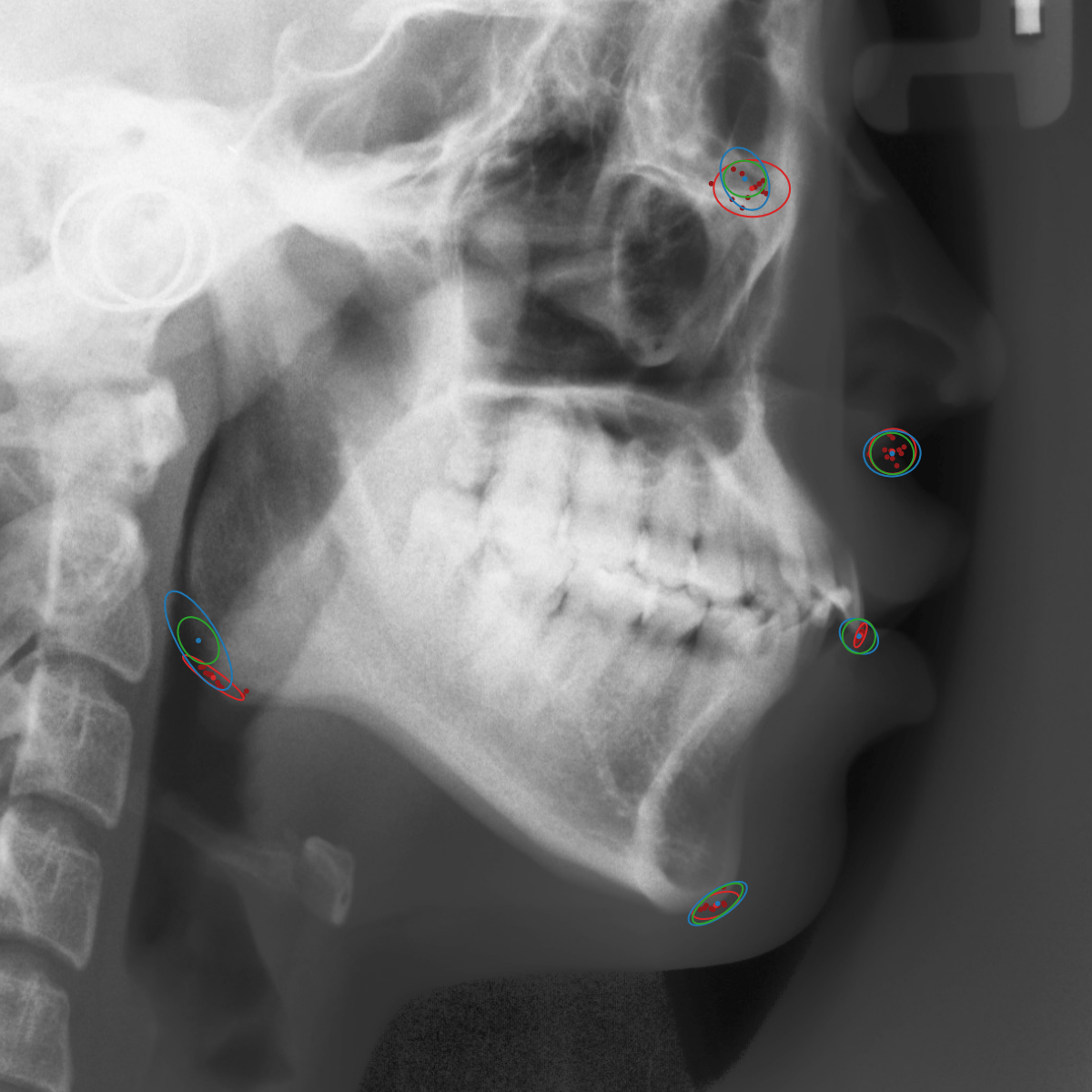} \\
\end{tabular}
\caption{
Qualitative results of the proposed $\netdirallfit$ method for selected samples 
of five depicted landmarks (bottom row).
The point annotations per landmark represent the annotations of the 11 observers (dark red), their mean (red) and the landmark prediction of our model (blue).
The distributions are represented as ellipses with semi-axes defined as $3 \sigmamajor$ and $3 \sigmaminor$ rotated by $\theta$, which are obtained by fitting a Gaussian function to the 11 annotations (red), to the predicted heatmaps $\predictioncovariance$-fit (blue) as well as to the target heatmap parameters $\targetcovariance$-target (green).
}
\label{fig:results_with_covariances}
\end{figure}

Qualitative results of the landmark predictions as well as the predicted distributions for each landmark are presented in Fig.~\ref{fig:results_with_covariances} for selected images.
The visualized points are the 11 inter-observer annotations (dark red), their mean position (red) and the landmark predictions of our model (blue).
The red ellipses represent the distributions fitted to the 11 annotations.
The \textit{sample-based uncertainties} and the \textit{dataset-based uncertainties} are represented as blue and green ellipses, respectively.
For the depicted images, it can be seen that most landmark predictions closely match to the mean of the 11 annotations, while the predicted uncertainties are similar to the annotation distributions.
Discrepancy between the inter-observer annotations and uncertainties predicted by our method were primarily caused by landmarks that where difficult to recognize in a given image (landmark 2, bottom left and landmark 4, bottom middle) or by landmarks with high ambiguity (e.g., landmark 3, bottom middle and landmark 2 bottom right).
%
Nevertheless, the predicted \textit{sample-based uncertainties} are similar to the annotation distributions even when the \textit{dataset-based uncertainty} of a landmark is too restrictive as, e.g., for landmark 2 (top middle), where both mandibles are visible due to the head being tilted.

\subsection{Ablation Study}
\label{sec:ablation}

\begin{table}[t]
\caption{
Gaussian parameters of the annotation distributions (Ann.), for target and fitted heatmaps for several strategies (\anisoaniso, \isoaniso, \fixediso), as well as for methods using MCD (\mcdpointfit,max, \mcdhmeananiso).
The product $\sigmamajor \cdot \sigmaminor$, the ratio $\sigmamajor~:~\sigmaminor$ and the rotation angle $\theta$ of the respective distributions are shown.
Additionally, the PE for each method is given.
The mean and standard deviation is shown where applicable.
For landmarks with low anisotropy and less indicative rotation ($L_3$, $L_4$, and $L_5$), the rotation angles are shown in gray.
}
\begin{center}
\scriptsize
\begin{tabular}{ cc|c|c|c|c|c }
\cline{1-7}
\cline{1-7}

\multicolumn{1}{c}{Method} & Metric & \multicolumn{1}{c}{$L_1$} & \multicolumn{1}{c}{$L_2$} & \multicolumn{1}{c}{$L_3$} & \multicolumn{1}{c}{$L_4$} & \multicolumn{1}{c}{$L_5$} \\

\cline{1-7}

\multirow{3}{*}{\rotatebox[origin=l]{90}{\parbox{2.5em}{\centering\mbox{Ann.}}}}
& $\sigmamajor:\sigmaminor$ & 2.57 & 2.12 & 1.52 & 1.11 & 1.46 \\
& $\sigmamajor \cdot \sigmaminor$ & 0.37 & 2.73 & 2.34 & 0.26 & 1.13 \\
& $\theta$ (in degree) & 39.33 & -44.30 & \gray{20.97} & \gray{-60.34} & \gray{-13.98} \\

\hline
\hline

\multirow{3}{*}{\rotatebox[origin=l]{90}{\parbox{2.5em}{\centering\mbox{$\predictioncovariance$-fit}}}}
& \multirow{1}{*}{$\sigmamajor:\sigmaminor$} & 2.80 $\pm$ 0.25 & 1.98 $\pm$ 0.47 & 1.34 $\pm$ 0.30 & 1.12 $\pm$ 0.12 & 1.20 $\pm$ 0.17 \\
& \multirow{1}{*}{$\sigmamajor \cdot \sigmaminor$} & 0.63 $\pm$ 0.07 & 1.40 $\pm$ 0.35 & 0.86 $\pm$ 0.18 & 0.44 $\pm$ 0.08 & 0.98 $\pm$ 0.24 \\
& \multirow{1}{*}{$\theta$ (in degree)} & 37.48 $\pm$ 2.72 & -56.62 $\pm$ 14.82 & \gray{-0.88 $\pm$ 35.90} & \gray{1.25 $\pm$ 57.68} & \gray{-30.52 $\pm$ 36.00} \\
\cline{2-7}
\multirow{4}{*}{\rotatebox[origin=l]{90}{\parbox{3em}{\centering\mbox{\vphantom{$\predictioncovariance$}$\covariance$-tar.,}}}}
& \multirow{1}{*}{$\sigmamajor:\sigmaminor$} & 2.63 & 1.51 & 1.17 & 1.03 & 1.05 \\
& \multirow{1}{*}{$\sigmamajor \cdot \sigmaminor$} & 0.48 & 0.62 & 0.52 & 0.40 & 0.62 \\
& \multirow{1}{*}{$\theta$ (in degree)} & 36.79 & -54.39 & \gray{-7.10} & \gray{-89.47} & \gray{-14.04} \\
\cline{2-7}
& PE (in mm) & 0.34 $\pm$ 0.25 & 1.31 $\pm$ 0.94 & 0.94 $\pm$ 0.67 & 0.41 $\pm$ 0.76 & 0.65 $\pm$ 0.47 \\

\cline{1-7}
\cline{1-7}

\multirow{3}{*}{\rotatebox[origin=l]{90}{\parbox{2.5em}{\centering\mbox{$\predictioncovariance$-fit}}}}
& \multirow{1}{*}{$\sigmamajor:\sigmaminor$} & 1.35 $\pm$ 0.14 & 1.70 $\pm$ 0.44 & 1.28 $\pm$ 0.22 & 1.13 $\pm$ 0.16 & 1.21 $\pm$ 0.19 \\
& \multirow{1}{*}{$\sigmamajor \cdot \sigmaminor$} & 0.76 $\pm$ 0.07 & 1.51 $\pm$ 0.38 & 0.88 $\pm$ 0.19 & 0.45 $\pm$ 0.11 & 1.02 $\pm$ 0.28 \\
& \multirow{1}{*}{$\theta$ (in degree)} & 37.16 $\pm$ 9.07 & -56.89 $\pm$ 25.12 & \gray{2.71 $\pm$ 55.33} & \gray{31.51 $\pm$ 54.57} & \gray{-40.34 $\pm$ 44.71} \\
\cline{2-7}
\multirow{4}{*}{\rotatebox[origin=l]{90}{\parbox{3em}{\centering\mbox{\vphantom{$\predictioncovariance$}$\sigma$-tar.,}}}}
& \multirow{1}{*}{$\sigmamajor:\sigmaminor$} & 1.00 & 1.00 & 1.00 & 1.00 & 1.00 \\
& \multirow{1}{*}{$\sigmamajor \cdot \sigmaminor$} & 0.52 & 0.63 & 0.52 & 0.39 & 0.62 \\
& \multirow{1}{*}{$\theta$ (in degree)} & - & - & \gray{-} & \gray{-} & \gray{-} \\
\cline{2-7}
& PE (in mm) & 0.35 $\pm$ 0.28 & 1.38 $\pm$ 0.99 & 1.00 $\pm$ 0.72 & 0.40 $\pm$ 0.75 & 0.65 $\pm$ 0.45 \\

\cline{1-7}
\cline{1-7}

\multirow{3}{*}{\rotatebox[origin=l]{90}{\parbox{2.5em}{\centering\mbox{$\predictioncovariance$-fit}}}}
& \multirow{1}{*}{$\sigmamajor:\sigmaminor$} & 1.29 $\pm$ 0.13 & 1.68 $\pm$ 0.39 & 1.27 $\pm$ 0.24 & 1.10 $\pm$ 0.09 & 1.17 $\pm$ 0.14 \\
& \multirow{1}{*}{$\sigmamajor \cdot \sigmaminor$} & 0.90 $\pm$ 0.09 & 1.46 $\pm$ 0.46 & 0.96 $\pm$ 0.20 & 0.70 $\pm$ 0.10 & 0.98 $\pm$ 0.22 \\
& \multirow{1}{*}{$\theta$ (in degree)} & 36.73 $\pm$ 11.78 & -56.92 $\pm$ 22.70 & \gray{-2.04 $\pm$ 55.45} & \gray{25.23 $\pm$ 49.33} & \gray{-34.09 $\pm$ 53.11} \\
\cline{2-7}
\multirow{4}{*}{\rotatebox[origin=l]{90}{\parbox{3em}{\centering\mbox{\vphantom{$\predictioncovariance$}$\sigma$=3,}}}}
& \multirow{1}{*}{$\sigmamajor:\sigmaminor$} & 1.00 & 1.00 & 1.00 & 1.00 & 1.00 \\
& \multirow{1}{*}{$\sigmamajor \cdot \sigmaminor$} & 0.66 & 0.66 & 0.66 & 0.66 & 0.66 \\
& \multirow{1}{*}{$\theta$ (in degree)} & - & - & \gray{-} & \gray{-} & \gray{-} \\
\cline{2-7}
& PE (in mm) & 0.35 $\pm$ 0.24 & 1.29 $\pm$ 0.92 & 0.97 $\pm$ 0.69 & 0.42 $\pm$ 0.77 & 0.64 $\pm$ 0.51 \\

\hline
\hline

\multirow{4}{*}{\rotatebox[origin=l]{90}{\parbox{3.5em}{\centering\mbox{\mcdpointfit}\\\mbox{max}}}}
& \multirow{1}{*}{$\sigmamajor:\sigmaminor$}
& 2.61 $\pm$ 0.86 & 4.84 $\pm$ 3.36 & 2.31 $\pm$ 1.55 & 1.73 $\pm$ 0.97 & 1.96 $\pm$ 1.11 \\
& \multirow{1}{*}{$\sigmamajor \cdot \sigmaminor$} & 0.01 $\pm$ 0.00 & 0.27 $\pm$ 0.55 & 0.06 $\pm$ 0.09 & 0.02 $\pm$ 0.11 & 0.02 $\pm$ 0.04 \\
& \multirow{1}{*}{$\theta$ (in degree)} & 33.89 $\pm$ 14.57 & -58.03 $\pm$ 28.26 & \gray{19.70 $\pm$ 50.21} & \gray{51.33 $\pm$ 32.61} & \gray{32.36 $\pm$ 44.69} \\
\cline{2-7}
& PE (in mm) & 0.35 $\pm$ 0.25 & 1.31 $\pm$ 0.87 & 0.99 $\pm$ 0.69 & 0.39 $\pm$ 0.66 & 0.65 $\pm$ 0.46 \\

\cline{1-7}
\cline{1-7}

\multirow{4}{*}{\rotatebox[origin=l]{90}{\parbox{3.5em}{\centering\mbox{\mcdhmean}\\\mbox{$\predictioncovariance$-fit}}}}
& \multirow{1}{*}{$\sigmamajor:\sigmaminor$}
& 1.31 $\pm$ 0.13 & 1.78 $\pm$ 0.47 & 1.27 $\pm$ 0.23 & 1.10 $\pm$ 0.10 & 1.17 $\pm$ 0.13 \\
& \multirow{1}{*}{$\sigmamajor \cdot \sigmaminor$} & 0.89 $\pm$ 0.08 & 1.48 $\pm$ 0.40 & 1.01 $\pm$ 0.22 & 0.72 $\pm$ 0.13 & 1.01 $\pm$ 0.22 \\
& \multirow{1}{*}{$\theta$ (in degree)} & 36.96 $\pm$ 9.61 & -56.74 $\pm$ 19.44 & \gray{4.26 $\pm$ 59.02} & \gray{25.17 $\pm$ 49.42} & \gray{-29.44 $\pm$ 48.50} \\
\cline{2-7}
& PE (in mm) & 0.34 $\pm$ 0.24 & 1.34 $\pm$ 1.04 & 0.97 $\pm$ 0.68 & 0.42 $\pm$ 0.77 & 0.65 $\pm$ 0.50 \\

\cline{1-7}
\cline{1-7}

\end{tabular}
\label{tb:ablation}
\end{center}
\end{table}

\newcommand{\smallerinterplotwidth}{0.16\textwidth}

\begin{figure}[t]
\centering
\begin{tabular}{ccc?{0.3mm}cc}
& \multicolumn{2}{c?{0.3mm}}{$L_1$} & \multicolumn{2}{c}{$L_2$} \\

\rotatebox[origin=l]{90}{\hspace{1.5em}\vphantom{p}{Ann.}}
& \multicolumn{2}{c?{0.3mm}}{\includegraphics[width=\smallerinterplotwidth]{figures/skull_new_plots/mean_all/anisotropic/obs_scatter_0.pdf}}
& \multicolumn{2}{c}{\includegraphics[width=\smallerinterplotwidth]{figures/skull_new_plots/mean_all/anisotropic/obs_scatter_1.pdf} } \\

\hdashline



\rotatebox[origin=l]{90}{\hspace{0.0em}\vphantom{p}{$\targetcovariance$-tar.,$\predictioncovariance$-fit}}
& \includegraphics[width=\smallerinterplotwidth]{figures/skull_new_plots/mean_all/anisotropic/pred_fit_0.pdf} 
& \includegraphics[width=\smallerinterplotwidth]{figures/skull_new_plots/mean_all/anisotropic/trained_param_0.pdf}
& \includegraphics[width=\smallerinterplotwidth]{figures/skull_new_plots/mean_all/anisotropic/pred_fit_1.pdf} 
& \includegraphics[width=\smallerinterplotwidth]{figures/skull_new_plots/mean_all/anisotropic/trained_param_1.pdf} \\

\rotatebox[origin=l]{90}{\hspace{0.2em}\vphantom{p}{$\sigma$-tar.,$\predictioncovariance$-fit}}
& \includegraphics[width=\smallerinterplotwidth]{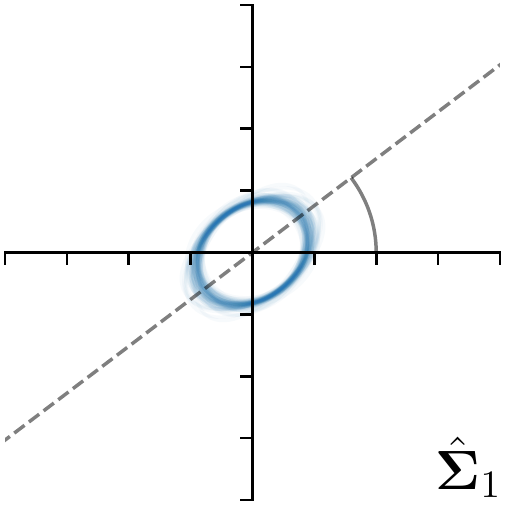} 
& \includegraphics[width=\smallerinterplotwidth]{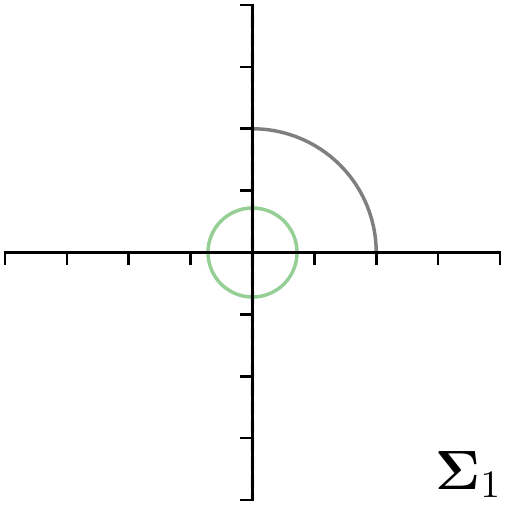}
& \includegraphics[width=\smallerinterplotwidth]{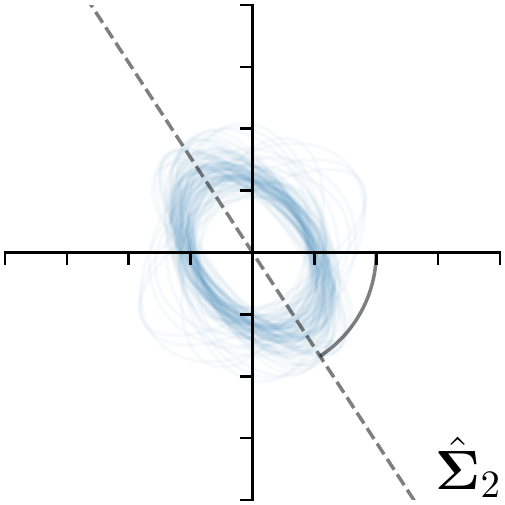} 
& \includegraphics[width=\smallerinterplotwidth]{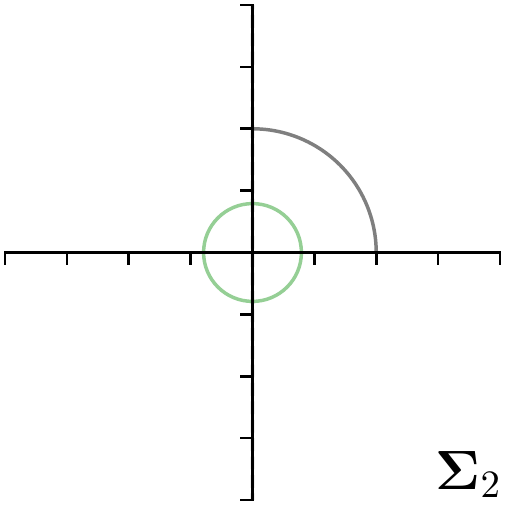} \\

\rotatebox[origin=l]{90}{\hspace{0.5em}\vphantom{p}{\fixedaniso}}
& \includegraphics[width=\smallerinterplotwidth]{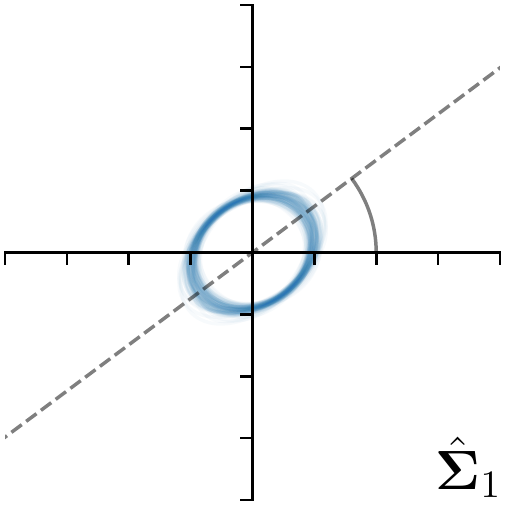} 
& \includegraphics[width=\smallerinterplotwidth]{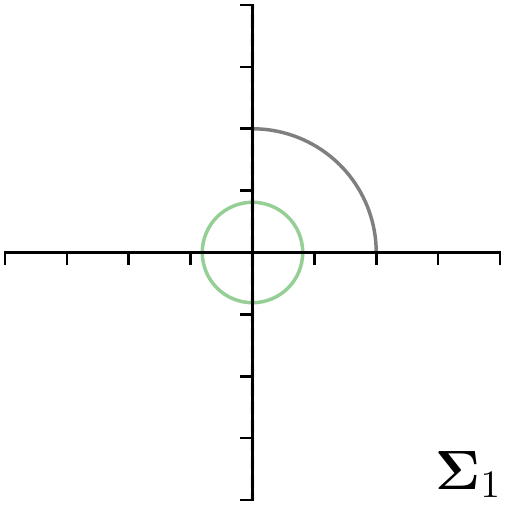}
& \includegraphics[width=\smallerinterplotwidth]{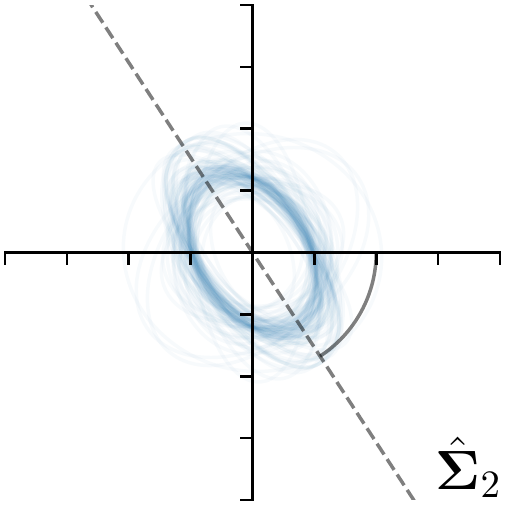} 
& \includegraphics[width=\smallerinterplotwidth]{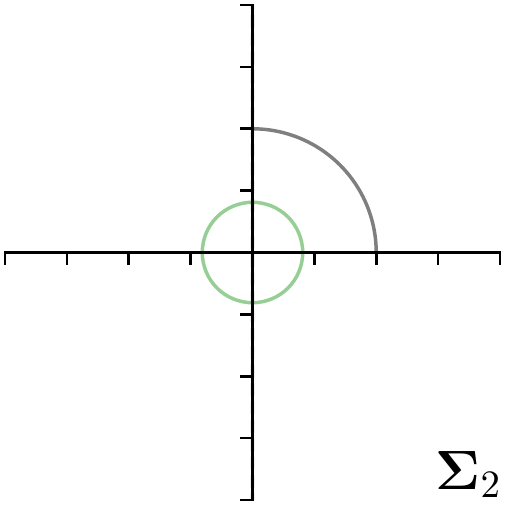} \\

\hdashline

\rotatebox[origin=l]{90}{\hspace{0.0em}\vphantom{p}{\mcdpointfit,max}}
& \includegraphics[width=\smallerinterplotwidth]{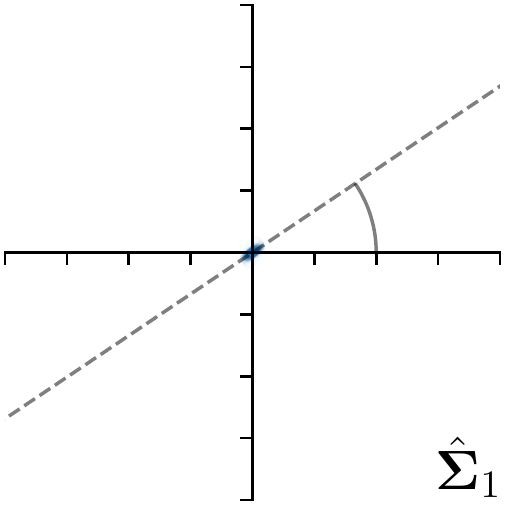} 
& \includegraphics[width=\smallerinterplotwidth]{figures/skull_new_plots/mean_all/fixed/trained_param_0.pdf}
& \includegraphics[width=\smallerinterplotwidth]{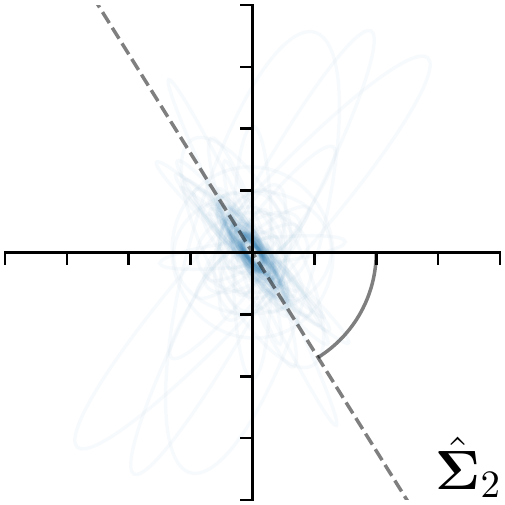} 
& \includegraphics[width=\smallerinterplotwidth]{figures/skull_new_plots/mean_all/fixed/trained_param_1.pdf} \\

\rotatebox[origin=l]{90}{\hspace{0.0em}\vphantom{p}{\mcdhmeananiso}}
& \includegraphics[width=\smallerinterplotwidth]{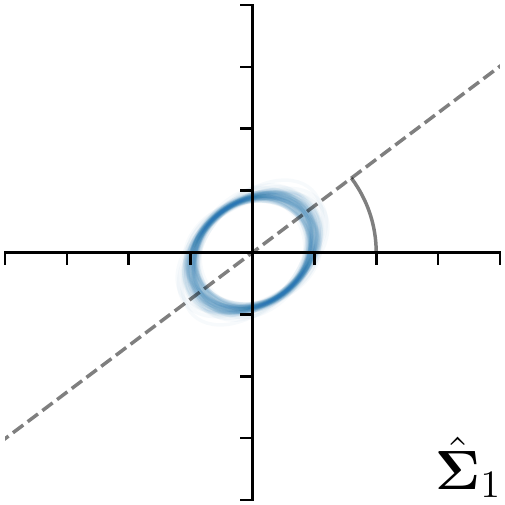} 
& \includegraphics[width=\smallerinterplotwidth]{figures/skull_new_plots/mean_all/fixed/trained_param_0.pdf}
& \includegraphics[width=\smallerinterplotwidth]{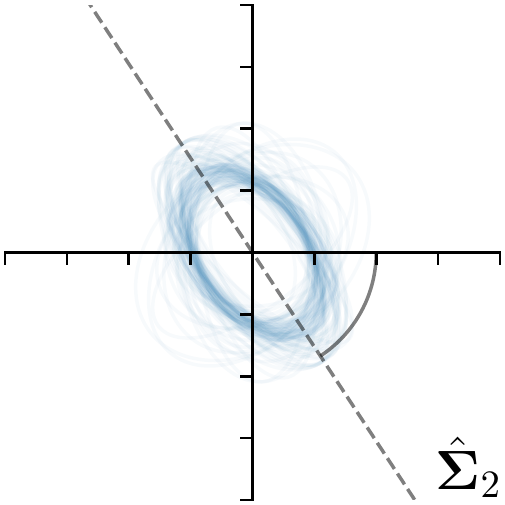} 
& \includegraphics[width=\smallerinterplotwidth]{figures/skull_new_plots/mean_all/fixed/trained_param_1.pdf} \\
\end{tabular}
\caption{
Distribution parameters for landmarks $L_1$ and $L_2$ visualized as ellipses for the annotation distributions (Ann.-Offsets), for target and fitted heatmaps for several strategies (\anisoaniso, \isoaniso, \fixediso), as well as for methods using MCD (\mcdpointfit,max, \mcdhmeananiso).
The visualizations of the individual plots are the same as in Fig.~\ref{fig:results:heatmap_plot_hand}.
}
\label{fig:ablation}
\end{figure}

Different strategies of using the Gaussian function to model both the target heatmap and the distribution of the heatmap prediction are investigated.
We distinguish methods based on whether the target heatmap used during training is defined as a Gaussian function with a fixed isotropic variance ($\sigma$=3), learned isotropic variance ($\sigma$-target) or learned anisotropic covariance ($\covariance$-target).
For the fixed isotropic variance, the value $\sigma=3.0$ was chosen, as it showed to have a good trade-off between PE and $\text{SDR}_r$.
As already defined, our main method uses the Gaussian function with anisotropic covariance during both training and inference, which we named $\anisoaniso$.
For comparison with state-of-the-art methods estimating uncertainty, we implemented two approaches using MCD as described in Sec.~\ref{sec:setup}.
The cephalogram dataset with our additional annotations was used for method comparison as well as the same visualizations and measures as in the previous Sec.~\ref{sec:inter}.
Results of these comparisons for two selected landmarks with the highest anisotropy are shown in Fig.~\ref{fig:ablation}, while the quantitative results for all five landmarks are shown in Table~\ref{tb:ablation}.

When comparing the localization accuracy in terms of PE, all different heatmap target and fitting strategies perform similarly well as can be seen in Table~\ref{tb:ablation}.
Similarly as in Sec.~\ref{sec:inter}, it can be seen that both target and predicted heatmaps have the tendency of underestimating the annotation distribution size $\sigmamajor \cdot \sigmaminor$ for landmarks having a large inter-observer variability.
When looking at the distributions modeled by the target heatmaps in Fig.~\ref{fig:ablation}, it can be observed that strategies using either fixed ($\sigma=3$) or learned isotropic $\sigma$ values ($\sigma$-target) have no capacity of modeling the annotation distribution.
Interestingly, when fitting anisotropic Gaussian functions to the heatmaps predicted by both of these strategies, the orientation angles $\theta$ of the inter-observer variability are successfully estimated. 
However, the ratio of $\sigmamajor : \sigmaminor$ in Table~\ref{tb:ablation} shows that our final model, which uses anisotropic Gaussian functions for both target heatmaps and prediction fitting ($\anisoaniso$), better captures the anisotropy of the annotator distributions.

Our method is also compared with the state-of-the-art method MCD for estimating uncertainties adapted to the landmark localization task, where the uncertainty is either computed as the variance of a set of stochastic forward predictions (\mcdpointfit,max) or estimated by fitting a Gaussian to an accumulated set of stochastic heatmaps (\mcdhmeananiso).
Since fixed target heatmaps were used to be consistent with related work, only the variances respectively the parameters of the fitted Gaussians are compared with the annotation distributions.
As can be seen in Table~\ref{tb:ablation}, there is no significant difference when comparing MCD to any of our strategies in terms of the PE.
However, $\mcdpointfit$,max is not capable of estimating the size of the annotation distributions $\sigmamajor \cdot \sigmaminor$, due to the predicted coordinates being very similar for all passes which leads to an underestimation.
Moreover, the ratio of $\sigmamajor : \sigmaminor$ is also unreliable for this method.
The method \mcdhmeananiso{} has a similar performance as our strategy \isoaniso, which was to be expected as the uncertainty is also estimated by fitting a Gaussian.

\subsection{Prediction Uncertainty in Clinical Measurements}
\label{sec:practical}

\begin{figure}[t]
\centering
\begin{tabular}{c}
Classification Accuracy \\
\includegraphics[width=0.9\textwidth]{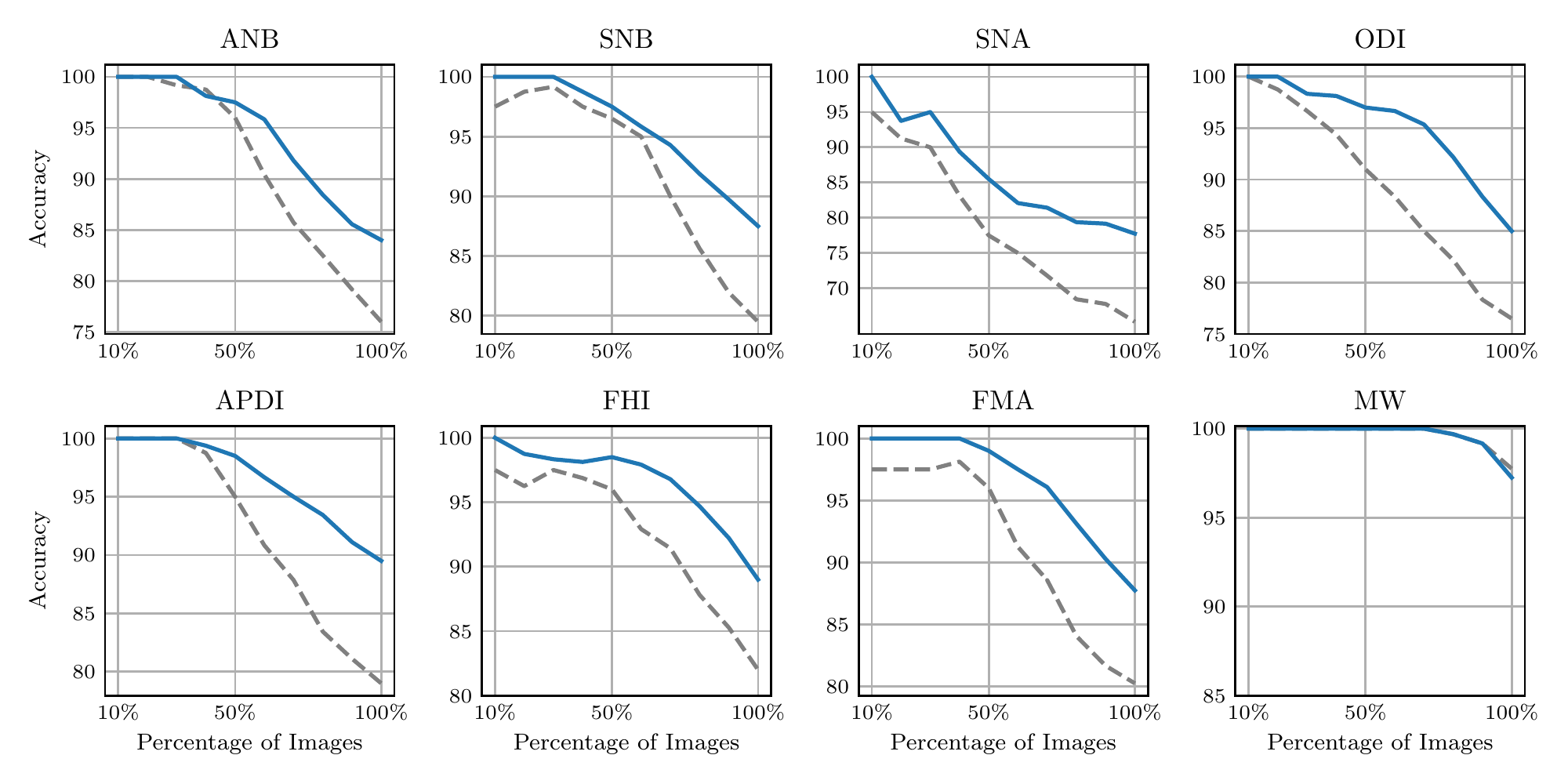} \\
\end{tabular}
\caption{
Classification accuracies of anatomical abnormalities using the cephalogram dataset, each shown in the respective subplot.
For each image and anatomical abnormality, the class probabilities as well as the corresponding classification uncertainties are calculated.
After sorting the images of the dataset by uncertainty, the plots show the classification accuracies (in percent) for different percentages of considered images.
For the same images, the dashed line represents the classification accuracy of the senior as compared to the junior annotator.
List of abbreviations:
ANB: Angle between A-point, nasion, B-point.
SNB: Angle between sella, nasion, B-point.
SNA: Angle between sella, nasion, A-point.
ODI: Overbite Depth Indicator.
APDI: Anteroposterior Dysplasia Indicator.
FHI: Facial Height Index.
FMA: Frankfort Mandibular Angle.
MW: Modified Wits Appraisal.
More detailed definitions of the classes are given in \cite{Wang2016} and \cite{Lindner2016}.
}
\label{fig:practical}
\end{figure}


In clinical practice utilizing computer aided diagnosis, often only the most probable landmark prediction is used, while the uncertainty distribution of the prediction is ignored leading to overconfidence in calculated clinical measurements.
To show the importance of considering the annotation uncertainty of the dataset used for training the machine learning model, we evaluate the influence of the predicted landmark localization uncertainty to the classification accuracy of clinical measurements for orthodontic analysis and treatment planning.
For this experiment, the same eight measurements that classify anatomical abnormalities as described in \cite{Wang2016} and listed in Fig.~\ref{fig:practical} are used.

To show the correlation of the predicted landmark localization uncertainty with the classification accuracy, we first calculated how the classification prediction is affected by the predicted landmark distributions.
Thus, for each patient and measurement, 10,000 points are sampled from each landmark distribution predicted by our $\anisoaniso$ method trained using the junior annotations to calculate 10,000 classification outcomes in the four-fold cross-validation setup (\textit{CV jun.}) that we used before in Sec.~\ref{sec:localization_experiments}.
For each patient and measurement, the entropy of the class probabilities is used to obtain the classification uncertainty, while the classification accuracy is calculated as the agreement of the most probable class prediction with the junior annotation.


To verify that such a calculated entropy is a valid measure of uncertainty, we compare the classification uncertainty with the classification accuracy, for each clinical measurement independently.
Fig.~\ref{fig:practical} visualizes the average accuracy that is calculated by increasing the number of considered patients, ordered from lower to higher uncertainty.
For all clinical measurements, the blue line in Fig.~\ref{fig:practical} shows that the average classification accuracy is decreasing when the number of images with higher classification uncertainty is increasing.
Thus, our experiment shows that the predicted landmark localization uncertainty can also be used to model classification uncertainty influencing the classification accuracy.

Additionally, the dashed line in Fig.~\ref{fig:practical} shows the classification accuracy of the senior as compared to the junior annotator, for the same order of images as used to generate the blue line.
When comparing the dashed with the blue line, it can be seen that the classification agreement of the senior to the junior is in general lower as compared to our method, which can be explained by our model being trained on the junior annotation.
However, it is more interesting to see that by ordering the images according to the uncertainty of our model, the accuracy of the senior's classification compared to the junior's is also decreasing.
An interesting extension would be to estimate the optimal heatmap size directly by maximizing the classification accuracy, however, this would go beyond the scope of this work focusing on landmark localization.
Nevertheless, the experiment in this subsection further confirms that our method was able to identify samples with higher ambiguities in annotation.

\section{Conclusion}
\label{sec:conclusion}
In the state-of-the-art methods for landmark localization, only the most probable prediction is used to estimate the landmark location, without considering the location distribution.
In this work, it was shown that the heatmap prediction can be used to model the location distribution reflecting the aleatoric uncertainty coming from ambiguous annotations. 
By learning anisotropic Gaussian parameters of the target heatmaps during optimization, our method is able to model annotation ambiguities of the entire training dataset.
Furthermore, by fitting anisotropic Gaussian functions to the predicted heatmaps during inference, our method is also able to model the prediction uncertainty of individual samples.
Additionally to the state-of-the-art results on both evaluated datasets, i.e., radiographs of left hands and lateral cephalograms, the results 
show that the size of the Gaussian function modeling the target heatmap as well as the size of the Gaussian function fitted to the predicted heatmap correlate with the localization accuracy.
Finally, the experiment on the cephalogram dataset annotated by 11 observers has shown that our method successfully models the inter-observer variability.
By measuring how the classification of anatomical abnormalities in lateral cephalograms is influenced by the predicted location uncertainty, we have shown the importance of integrating the uncertainty into decision making to identify overconfident predictions, thus, potentially improving patient treatment.






\acks{This work was supported by the Austrian Research Promotion Agency (FFG): 871262.}

%
\ethics{The work follows appropriate ethical standards in conducting research and writing the manuscript, following all applicable laws and regulations regarding treatment of animals or human subjects.}


\coi{We declare we don't have conflicts of interest.}

\bibliography{library}




\end{document}